\documentclass[runningheads]{llncs}
\usepackage[utf8]{inputenc}
\usepackage{float}

\usepackage[T1]{fontenc}
\usepackage{graphicx}
\usepackage[table]{xcolor} % <--- solo aquí

\usepackage{bera}
\usepackage{listings}
\usepackage{enumitem}
\usepackage{pgfplots}
\usepackage{booktabs}
\pgfplotsset{compat=1.18}
\usepgfplotslibrary{polar}
\usepackage{tcolorbox}
\usepackage{subcaption}

\newtcbox{\tagLLM}{on line, colback=purple!15!white, colframe=purple!70!black,
  fontupper=\bfseries\footnotesize, boxsep=2pt,
  left=6pt, right=6pt, top=3pt, bottom=3pt, arc=5pt, boxrule=0.7pt}

\newtcbox{\tagLoss}{on line, colback=orange!15!white, colframe=orange!80!black,
  fontupper=\bfseries\footnotesize, boxsep=2pt,
  left=6pt, right=6pt, top=3pt, bottom=3pt, arc=5pt, boxrule=0.7pt}

\newtcbox{\tagTechniques}{on line, colback=cyan!10!white, colframe=cyan!70!black,
  fontupper=\bfseries\footnotesize, boxsep=2pt,
  left=6pt, right=6pt, top=3pt, bottom=3pt, arc=5pt, boxrule=0.7pt}

\newtcbox{\tagModelType}{on line, colback=green!15!white, colframe=green!60!black,
  fontupper=\bfseries\footnotesize, boxsep=2pt,
  left=6pt, right=6pt, top=3pt, bottom=3pt, arc=5pt, boxrule=0.7pt}

\newtcbox{\tagBias}{on line,
  colback=blue!15!white, colframe=blue!50!black,
  fontupper=\bfseries\footnotesize, boxsep=2pt,
  left=6pt, right=6pt, top=3pt, bottom=3pt, arc=5pt, boxrule=0.7pt}

\newtcbox{\tagExternalData}{on line,
  colback=brown!10!white, colframe=brown!60!black,
  fontupper=\bfseries\footnotesize, boxsep=2pt,
  left=6pt, right=6pt, top=3pt, bottom=3pt, arc=5pt, boxrule=0.7pt}
  
\newtcbox{\tagLabelModelType}{on line, colback=green!15!white, colframe=green!60!black,
  fontupper=\small, boxsep=1pt, left=2pt, right=2pt, top=1pt, bottom=1pt, arc=4pt, boxrule=0.5pt}

\newtcbox{\tagLabelLoss}{on line, colback=orange!15!white, colframe=orange!60!black,
  fontupper=\small, boxsep=1pt, left=2pt, right=2pt, top=1pt, bottom=1pt, arc=4pt, boxrule=0.5pt}

\newtcbox{\tagLabelTechniques}{on line, colback=cyan!10!white, colframe=cyan!70!black,
  fontupper=\small, boxsep=1pt, left=2pt, right=2pt, top=1pt, bottom=1pt, arc=4pt, boxrule=0.5pt}

\newtcbox{\tagLabelLLM}{on line, colback=purple!15!white, colframe=purple!70!black,
  fontupper=\small, boxsep=1pt, left=2pt, right=2pt, top=1pt, bottom=1pt, arc=4pt, boxrule=0.5pt}

\newtcbox{\tagLabelBias}{on line, colback=blue!15!white, colframe=blue!50!black,
  fontupper=\small, boxsep=1pt, left=2pt, right=2pt, top=1pt, bottom=1pt, arc=4pt, boxrule=0.5pt}

\newtcbox{\tagLabelExternalData}{on line,
  colback=brown!10!white, colframe=brown!60!black,
  fontupper=\small,  boxsep=1pt, left=2pt, right=2pt, top=1pt, bottom=1pt, arc=4pt, boxrule=0.5pt}
  
\colorlet{punct}{red!60!black}
\definecolor{background}{HTML}{EEEEEE}
\definecolor{delim}{RGB}{20,105,176}
\colorlet{numb}{magenta!60!black}

\lstdefinelanguage{json}{
    basicstyle=\normalfont\ttfamily,
    numbers=left,
    numberstyle=\scriptsize,
    stepnumber=1,
    numbersep=8pt,
    showstringspaces=false,
    breaklines=true,
    frame=lines,
    backgroundcolor=\color{background},
    literate=
     *{0}{{{\color{numb}0}}}{1}
      {1}{{{\color{numb}1}}}{1}
      {2}{{{\color{numb}2}}}{1}
      {3}{{{\color{numb}3}}}{1}
      {4}{{{\color{numb}4}}}{1}
      {5}{{{\color{numb}5}}}{1}
      {6}{{{\color{numb}6}}}{1}
      {7}{{{\color{numb}7}}}{1}
      {8}{{{\color{numb}8}}}{1}
      {9}{{{\color{numb}9}}}{1}
      {:}{{{\color{punct}{:}}}}{1}
      {,}{{{\color{punct}{,}}}}{1}
      {\{}{{{\color{delim}{\{}}}}{1}
      {\}}{{{\color{delim}{\}}}}}{1}
      {[}{{{\color{delim}{[}}}}{1}
      {]}{{{\color{delim}{]}}}}{1},
}

\usepackage{booktabs}
\usepackage{siunitx}
\usepackage[table]{xcolor}

\definecolor{gold}{rgb}{1.0, 0.84, 0}
\definecolor{silver}{rgb}{0.8, 0.8, 0.8}
\definecolor{bronze}{rgb}{0.8, 0.5, 0.2}

\begin{document}
\title{Overview of the TalentCLEF 2025: Skill and Job Title Intelligence for Human Capital Management}
%An overview of the BIOASQ large-scale biomedical semantic indexing and question answering competition

%
\titlerunning{TalentCLEF 2025}

% Paula Estrella
% Rabih Zbib 0000-0002-7140-3048
% Laura García-Sardiña 0000-0003-4592-8884
% Luis Gasco 0000-0002-4976-9879
% Alvaro Rodrigo 0000-0002-6331-4117
% Hermenegildo Fabregat 0000-0001-9820-2150
% Daniel Deniz 0000-0002-0313-2127
\author{Luis Gasco\inst{1}\orcidID{0000-0002-4976-9879} \and
Hermenegildo Fabregat\inst{1,2}\orcidID{0000-0001-9820-2150} \and
Laura García-Sardiña\inst{1}\orcidID{0000-0003-4592-8884} \and
Paula Estrella\inst{1}\and
Daniel Deniz\inst{1}\orcidID{0000-0002-0313-2127} \and
Alvaro Rodrigo\inst{2}\orcidID{0000-0002-6331-4117} \and
Rabih Zbib\inst{1}\orcidID{0000-0002-7140-3048}
}

%
% Luis Gasco 0000-0002-4976-9879

\authorrunning{L. Gasco et al.}
% GILDO UNED Y AVATURE. 
\institute{Avature Machine Learning, Spain \\
\email{machinelearning@avature.net} \and
NLP \& IR Group at UNED, Madrid, Spain
}

\maketitle              % typeset the header of the contribution
\pagestyle{empty}
\begin{abstract}

Advances in natural language processing and large language models are driving a major transformation in Human Capital Management, with a growing interest in building smart systems based on language technologies for talent acquisition, upskilling strategies, and workforce planning. However, the adoption and progress of these technologies critically depend on the development of reliable and fair models, properly evaluated on public data and open benchmarks, which have so far been unavailable in this domain.

To address this gap, we present TalentCLEF 2025, the first evaluation campaign focused on skill and job title intelligence. The lab consists of two tasks: Task A -  Multilingual Job Title Matching, covering English, Spanish, German, and Chinese; and Task B - Job Title-Based Skill Prediction, in English. Both corpora were built from real job applications, carefully anonymized, and manually annotated to reflect the complexity and diversity of real-world labor market data, including linguistic variability and gender-marked expressions. %The evaluation scenarios included settings of monolingual, cross-lingual, and gender bias.
The evaluations included monolingual and cross-lingual scenarios and covered the evaluation of gender bias.

TalentCLEF attracted 76 registered teams with more than 280 submissions. Most systems relied on information retrieval techniques built with multilingual encoder-based models fine-tuned with contrastive learning, and several of them incorporated large language models for data augmentation or re-ranking. The results show that the training strategies have a larger effect than the size of the model alone. TalentCLEF provides the first public benchmark in this field and encourages the development of robust, fair, and transferable language technologies for the labor market.

\end{abstract}
\clearpage

\section{Introduction}
% Why this task is relevant. 
% Take the justification of the proposal as idea.
% LUIS

% Start con un dato
The landscape of the global labor market is undergoing a profound transformation driven by rapid technological advancements. Recent studies suggest that by 2030, approximately 70\% of the skills required for today's professions will have changed, and a significant proportion of the workforce will be employed in occupations that did not exist at the beginning of the twenty-first century~\cite{linkedin2025workchange}.

% Explicar el por qué del anterior stament.
This change is not accidental, but rather a consequence of the technological development experienced in recent years, which has transformed our understanding of existing job roles. Digital transformation, automation, and the rise of artificial intelligence are redefining the skills needed in the workforce. These advancements have not only automated tasks that were once routine and created entirely new roles, but have also enabled remote work, which is eliminating the geographical boundaries of employment and sourcing talent.

% Challenges de las empresas y organizaciones en el cambio. 
In this transition, the challenge of adaptation affects both individuals and organizations. On the one hand, professionals face increasing uncertainty about which skills will be relevant in the near future, making the need for reskilling more pressing than ever. On the other hand, companies face the task of identifying and attracting talent equipped with these emerging capabilities~\cite{manpower2024talentshortage}.

% Solucion al cambio y problemas es las language technologies aplicadas a este campo
All of these challenges have encouraged the adoption of language technologies applied to Human Capital Management (HCM). These technologies are used primarily for talent acquisition, helping organizations match candidates to job positions based on their previous roles and skills. They also support onboarding and training by enabling the creation of customized learning pathways adapted for each employee. Additionally, they play an increasingly important role in strategic workforce planning, as they allow companies to anticipate market trends and future skill requirements.

% El uso de esas tecnologías presenta retos.
However, the development and use of these technologies present important challenges~\cite{gasco2025talentclef}. First, \textbf{multilingualism} is still a barrier. In a global market where companies operate in multiple countries and languages, the ability to process information in different languages is essential for effective talent management. Second, it is crucial to ensure that the developed systems are \textbf{fair and unbiased}, especially when they are used to make decisions about hiring, training, or granting promotions. We need to identify and reduce algorithmic biases that could negatively impact certain groups, such as those defined by gender or ethnicity. Finally, \textbf{adaptability} is another challenge, as the importance of skills in candidates can differ widely between industries, or across time. This means that we need flexible systems that can adapt to the specific needs of different sectors.

% Se ha dado respuesta a los retos, fomentadno el desarrollo. 
In response to these challenges, the research community has focused on developing NLP techniques for HCM. Recently, there has been an increase in research activities in areas such as job titles and skill extraction and normalization into standard terminologies such as ESCO or O*NET ~\cite{lavi2021consultantbert,retyk2023r,retyk2024melo,senger2024deep,zhang2022skill,zhang2022skillspan,zhang2023escoxlm}, job-skill relations~\cite{fabregat2024inductive,giabelli2021skills2job}, and job title matching ~\cite{decorte2021jobbert,deniz2024combined,zbib2022learning}. This interest has fueled the organization of several research workshops, such as \textit{NLP4HR}~\cite{hruschka2024proceedings} or \textit{RecSys in HR}~\cite{bogers2024fourth}, among others~\cite{zhu20245th,ai4hrpes2023}, which have contributed to the emergence and consolidation of a research community in this area. Despite these advances, the absence of standardized benchmarks continues to hinder progress. Most research relies on proprietary datasets due to privacy constraints and, when public datasets exist, they often suffer from shortcomings like the lack of transparency regarding labeling criteria and missing evaluation scripts for comparison when trying to advance state-of-the-art. 

% Sin embargo aún faltan evaluation campaigns, que es lo que presentamos en el paper.
To help address these limitations and promote NLP research in the labor domain, this paper presents the results of the TalentCLEF 2025 lab, the first evaluation campaign in Human Capital Management. TalentCLEF introduces a series of challenges to evaluate important tasks in the HCM area, such as job title matching. The data used attempt to better capture the real-word complexities of the task by including terminological variability, multilingualism, and gender-marked variations. Inspired by successful community initiatives in other areas such as BioASQ~\cite{nentidis2022overview} and BioCreative~\cite{miranda2023overview}, TalentCLEF aims to drive the creation of open and robust evaluation resources and to encourage the development of more fair, more transferable, and more adaptable solutions for talent management systems.

\section{Overview of the Tasks}
% Primero, presento las dos tareas abordadas en TalentCLEF2025
The first edition of TalentCLEF is focused on the development and evaluation of NLP models for key applications in human capital management: (i) identify suitable candidates for job positions based on professional experience and skills, (ii) implement upskilling programs to foster continuous employee development, and (iii) detect emerging skills and skill gaps within organizations. The TalentCLEF lab consists of two tasks tailored to these goals: Task A - Multilingual Job Title Matching and Task B - Job Title-Based Skill Prediction.

\subsection{Task A: Multilingual Job Title Matching}
Job title matching is a fundamental task in NLP for human resources. In talent sourcing, systems must recognize when different job titles, even across multiple languages, refer to the same underlying role, a capability that is essential for building accurate and fair job-candidate recommender systems. The aim of Task A is to develop systems that can identify and rank the job titles most similar to a given one. For each job title, participating teams are required to generate a ranked list of similar job titles from a specified knowledge base. 

% About the data
In the task, we provide a multilingual dataset for job title matching, covering English, Spanish, German, and Chinese. The dataset consists of three partitions: a training set, available only for German, English, and Spanish, and is automatically generated as pairs of related job titles extracted from the ESCO taxonomy; and development and test sets, both created and annotated manually as described in Section~\ref{corpus_overview}. Table~\ref{tab:corpus_stats} summarizes key statistics of the corpus. Data for Task A were made available to participating teams via Zenodo\footnote{Task A corpus: \url{https://doi.org/10.5281/zenodo.14002665}}.

\begin{table}[ht]
\centering
\caption{Summary statistics of the development and test sets of Task A by language (en: English, es: Spanish, de: German, zh: Chinese).}
\label{tab:corpus_stats}
\begin{tabular}{lcccccccc}
\toprule
 & \multicolumn{4}{c}{\textbf{Dev}} & \multicolumn{4}{c}{\textbf{Test}} \\
\cmidrule(lr){2-5} \cmidrule(lr){6-9}
 & en & es & de & zh & en & es & de & zh  \\
\midrule
\# Queries           & 105 & 185 & 203 & 103 & 116 & 191 & 226 & 116 \\
\# Corpus Elements   & 2,619 & 4,661 & 4,729 & 2,513 & 769 & 1,231 & 1,509 & 769 \\
\ Avg. relevant items per query & 23.0 & 41.0 & 41.5 & 22.5 & 32.9 & 57.9 & 65.1 & 32.9 \\
\bottomrule

\end{tabular}
\end{table}

% Move this to evlauation section.
% The evaluation has been carried out on the Codabench platform\footnote{\url{https://www.codabench.org/competitions/5842/}}, where participants have submitted their results. Three main evaluation scenarios are considered: (i) a monolingual evaluation, where the system must identify similar job titles within the same language; (ii) a cross-lingual evaluation, where the system is required to recognize and match equivalent job titles across different languages; and (iii) a gender-based evaluation, which assesses the sensitivity of models to the grammatical gender of job titles, in order to detect possible biases in models. The official evaluation metric of the task is Mean Average Precision (MAP), but other metrics such as Rank-Biased Overlap (RBO) are also used to assess gender bias.

\subsection{Task B: Job Title-Based Skill Prediction}
Recent professional reports underscore the growing importance of a skill-centric approach to talent management. In fact, more than half of Europe's workforce is estimated to require reskilling in the coming years due to technological innovations like AI and automation~\cite{smit2020future,di2023future}. This emerging need for new skills presents a big challenge for companies and employees in their sourcing and training. In this context, developing models that predict skills that align with specific job titles will be essential to help organizations identify current and future skill gaps, enable targeted upskilling and reskilling initiatives, and also to support workforce planning based on competencies. The aim of Task B is to develop systems that, given a set of skills from a knowledge base, identify and rank by relevance those that are relevant to a given job position. 

In this task, we provide a dataset in English for job title-based skill relevance\footnote{Task B corpus: \url{https://doi.org/10.5281/zenodo.14002665}}. The dataset is divided into three partitions: a training set, which has been automatically generated leveraging the job title to skill relevance information from ESCO; and development and test sets, manually annotated following the process described in Section~\ref{corpus_overview}. Table~\ref{tab:corpus_stats_b} shows the statistics of the corpus.

% HERE THE TABLE 
\begin{table}[ht]
\caption{Summary statistics of the Task B corpus for the development and test sets.}
\centering
\label{tab:corpus_stats_b}
\begin{tabular}{lcc}
\toprule
 & \textbf{Dev} & \textbf{Test} \\
\midrule
\# Queries         & 304 & 125 \\
\# Corpus Elements & 1,439 & 1,986 \\
\ Avg. relevant items per query &85.2 & 101.8 \\
\bottomrule
\end{tabular}
\end{table}

%HERE ABOUT THE EVALUATION:
%For evaluation, we set up a Codabench competition\footnote{\url{https://www.codabench.org/competitions/7059/}}, where participants could submit their results. We evaluated the highest-performing system based on MAP (Mean Average Precision).

\subsection{Evaluation} 
% Include here information about metrics, etc. Maybe move here the info about codabench, and evaluation type.
The official evaluation for both tasks was done using the Codabench platform, with specific competitions for Task A\footnote{\url{https://www.codabench.org/competitions/5842/}} and Task B\footnote{\url{https://www.codabench.org/competitions/7059/}}. This structure offered participants a unified interface for submitting results and accessing leaderboards, while also establishing an open benchmark for continuous evaluation after the end of the task.

Evaluation of Task A considered four different scenarios: (i) \textbf{Averaged monolingual evaluation}, where systems had to identify similar job titles within the same language in English, Spanish and German; (ii) \textbf{cross-lingual evaluation}, where systems had to match job titles in Spanish and German given English queries; (iii) \textbf{Chinese language track} allowed teams to voluntary submit predictions for Chinese job titles, even though no training data was provided for this language, challenging participants to find solutions for this setting; and (iv)  \textbf{gender-based evaluation}, to assess the sensitivity of models to the grammatical gender of job titles, in order to detect possible biases in models for both Spanish and German. On the other hand, Task B had a single evaluation scenario, focusing exclusively on English. In this task, systems were assessed on their ability to identify relevant skills for each job title using the provided dataset in English.

For both Task A and Task B, the primary evaluation metric is \textbf{Mean Average Precision (MAP)}, a standard metric for information retrieval tasks in NLP. MAP measures the ability of a system to return highly relevant items at the top of a ranked list, averaged across all queries. This metric is particularly suitable for evaluating systems designed to prioritize the most relevant results, making it a natural choice for both job title matching and skill prediction tasks.

% In addition to MAP, the gender-based evaluation scenario in Task A also used the \textbf{Rank Biased Overlap (RBO)} metric~\cite{wmz10:acmtois}. RBO compares how similar two ranked lists are, giving more importance to the top of the list. In our evaluation, RBO was used to measure how much a system’s ranking changed when job titles appeared in different gender forms. A high RBO score means the system gave stable rankings regardless of gender, helping us detect and quantify gender bias.

In addition to MAP, Task A also used the \textbf{Rank Biased Overlap (RBO)} metric to evaluate gender bias~\cite{wmz10:acmtois}. RBO is a metric that compares the similarity between two ranked lists, placing more weight on items near the top of each list. This reflects the intuition that top-ranked results matter more, which is especially relevant in real-world systems like search engines or recommendation models. Unlike other ranking metrics, RBO is robust to differences in the length and content of the inputs lists, and it can handle partial or non-overlapping lists.

RBO was used to measure how a system's output ranking changes when queries (job titles) are presented in different gendered forms (e.g., \textit{``enfermero''} versus \textit{``enfermera''}, \textit{``abogado''} versus \textit{``abogada''}). By comparing both rankings, RBO quantifies how much the change in wording affects the system's output. A high RBO score indicates that rankings remain consistent despite gender variation, suggesting low sensitivity to gendered language. In contrast, a low RBO score indicates that the gender dimension significantly influences the results i.e., an important signal for detecting and measuring gender bias in ranking models.

\section{Corpus Overview}\label{corpus_overview}

The data provided to participants as a training set was generated using the ESCO taxonomy, while the TalentCLEF 2025 development and test sets were developed using real, de-identified job application data to ensure both authenticity and compliance with privacy regulations. Each of them was derived from job offers and the resumes of their associated applicants, enabling the extraction of positive pairs. In other words, applicants were considered relevant as they had applied to the corresponding position.  

The datasets were tailored to the specific retrieval objectives of each task. In both cases, the query component was defined as the job offer title. For Task A, the corpus consisted of job titles extracted from applicant resumes, reflecting how candidates describe their professional roles. For Task B, the corpus was composed of skills automatically extracted from applicant CVs (curricula vitae), allowing for skill-based relevance modeling.

\begin{figure}[hbt!]
    \centering
    \includegraphics[width=1\textwidth]{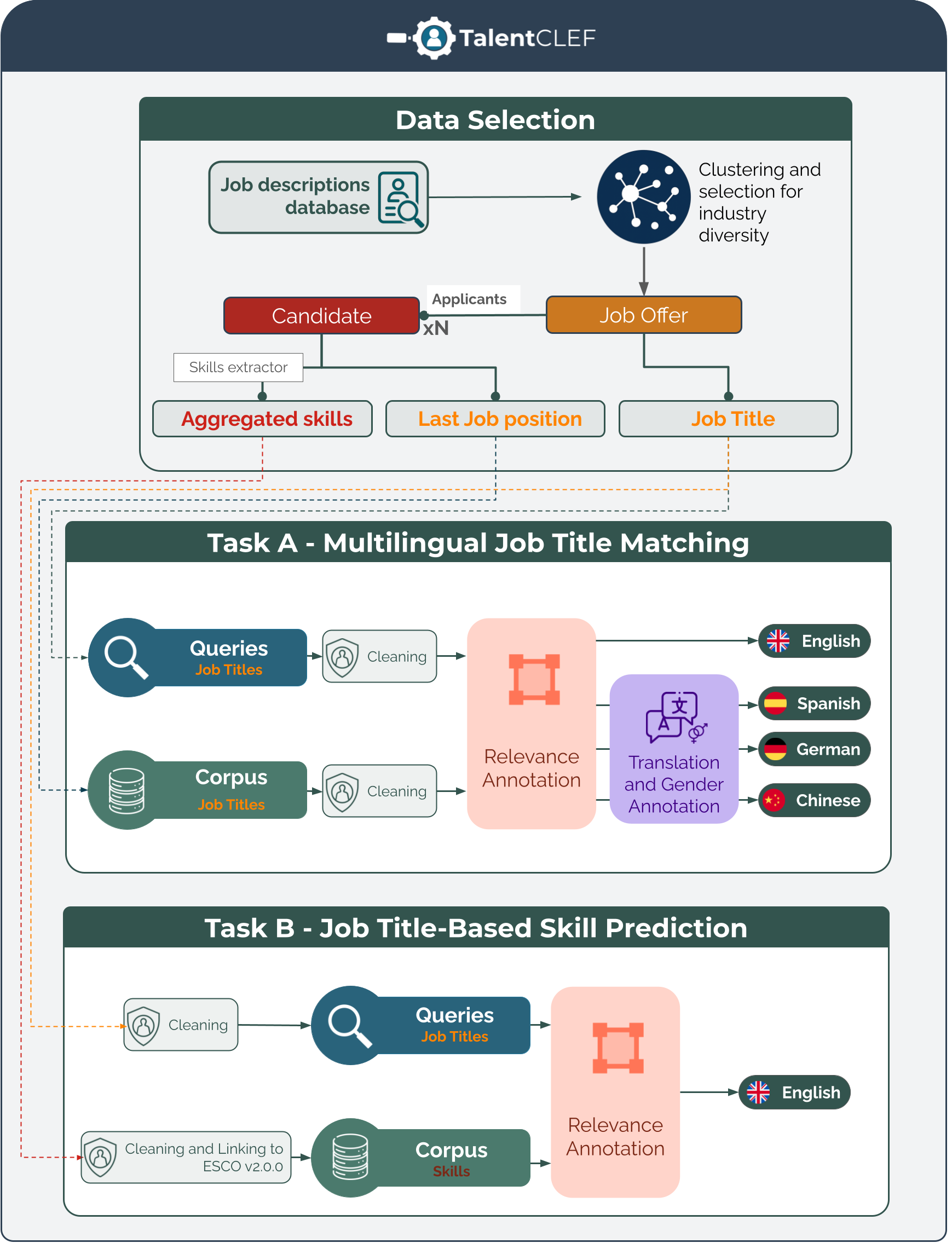}
    \caption{TalentCLEF 2025 - Corpus creation workflow.}
    \label{fig:dataset_creation}
\end{figure}

The process of constructing the datasets was designed to maximize diversity in terms of industry, role, language, and gender, while introducing the type of noise and heterogeneity typical of real-world recruitment pipelines.  The workflow used to create these resources is shown in Figure~\ref{fig:dataset_creation}, where the schema for data selection, creation and annotation of each corpus is represented in blocks.

\subsection{Data Selection}

The creation of the TalentCLEF 2025 corpus began with the collection of a large pool of job descriptions and their associated candidate applications from a real-world database. To ensure semantic and domain diversity, job offers were clustered using K-means and representative offers were sampled from different clusters to maximize coverage across industries and job roles.

For each selected offer, a fixed number ($xN$) of relevant candidate profiles was retrieved \textemdash 
 specifically, those who had actually applied to the position. From each candidate profile, both the most recent job title and a comprehensive set of skills were extracted using a high-recall skill extraction system designed to capture any potentially relevant skills for the job offer.

The result is a rich dataset where, for each job offer, we have: (1) the job title of the offer, (2) the most recent job titles of the candidates who applied for that offer, and (3) a group of candidate skills, potentially relevant to the targeted job offer.
%All data underwent de-identification and normalization prior to annotation, safeguarding personal information and ensuring consistency of format, while preserving naturally occurring linguistic variation and data noise to introduce realistic challenges for retrieval systems.

\subsection{Corpus Creation}
\subsubsection{Task A}

For Task A, queries correspond to cleaned titles of job offer, while corpus elements are the most recent job titles extracted from applicants. All job titles underwent a rigorous cleaning phase to remove sensitive or irrelevant information such as company names, locations, job codes, or other extraneous details (e.g., transforming ''\emph{CLOSED - Clinical Trial Manager – Remote}'' into ''\emph{Clinical Trial Manager}''). Annotators were instructed to flag titles that were incomplete, ambiguous, or not genuine job titles for exclusion from the dataset (e.g. ''\emph{Automotive, Engineering, Management, Manufacturing, QA - Quality Control}'' was considered a department rather than a job title).

A major goal was to maintain the realism of the data. Thus, while normalization removed sensitive content, it preserved natural variation, typographical errors, and incomplete entries. This presents retrieval systems with more realistic and noisy data, and better thus reflects the challenges of real-world talent matching.

Once the data had been cleaned, the annotation process consisting in two distinct stages was applied:
\begin{enumerate}
\item Annotators first judged the relevance of each candidate job title to the given job offer, following detailed guidelines. This often required extra research to clarify duties, industry context, and seniority, especially for ambiguous titles or acronyms such as ''\emph{OAE Evaluator}''\footnote{''\emph{OAE Evaluator}'' could be someone assessing otoacoustic emissions (OAEs), or someone specialized in evaluating educational qualifications (Office of Audit and Evaluation)}. Annotators used online sources to disambiguate and expand such cases when possible.
\item An extensive review phase then aligned annotation criteria, corrected over-annotation, and harmonized judgments. In total, 28.4\% of the initial matches were edited, most of them coming from a single annotator.
\end{enumerate}

After completion of the relevance annotation in English, the dataset was manually translated into Spanish, German, and Chinese by professional linguists. For Spanish and German, masculine, feminine, and neutral forms were generated where applicable. 

\subsubsection{Task B}
For Task B (Job Title-Based Skill Prediction), the same queries and candidate pools were used, but the corpus elements consisted of skill sets automatically extracted from candidate documents. These skill lists were curated to identify the most representative skills for each candidate based on their total frequency, we selected a total of 7,493 skills that appeared at least 90 times in the dataset. These skills were manually mapped to ESCO taxonomy v1.2.0 to ensure standardization in the lexical forms across occurrences of the same skill and to facilitate relevance modeling.

The relevance modeling was done by 4 annotators, who annotated the skills relevant to a given job title using the ESCO taxonomy as a reference to find the most relevant skills for each job offer. The process involved multiple stages, including a quality control initial stage and a complete final review of the output to ensure the accuracy and consistency of the gold standard. The test set contained 2,400 annotations and about a third was edited during review. Both development and test sets were annotated by the same pool of linguists and independently reviewed by an expert.

\section{Participants}
% General introduction to the participation. Include some stats as people registered from acadeima/industry; and country
\subsection{Task A}\label{subsec:participants_taskA}
% Number of registered people. Number of submissions. 
% La tarea ha tenido XX registrados en la Tarea, de los cuales 9 equipos han participado enviado submissions. En total, entre development y test set se han realizado un total de XXXX runs. 
% LA tabla muestra un overview d elas metodologías utilizadas. La etiqueta XXXXX se refuere al tipo de modelos utilizados, Techniques representa el tipo d emetodología utilizada para reoslver la tarea, LabelLoss representa el tipo de los utilizada si se ha llvado un ajuste, la etiqueta LLM-related presenta el uso que se le ha dado a los LLMs generativos, si es el caso, y por último, la etiqueta LabelBias es si se ha llevado a cabo agún tipo de gestión para address el gender bias. 

A total of 66 teams registered for the task, with 16 teams submitting at least one run in the evaluation phase, and 12 deciding to be part of the benchmark. Across both the development and the test sets, participants contributed a total of 196 submissions, showing a strong engagement with the challenge. Table~\ref{tab:participant_task_a} provides an overview of the methods used by the teams. \tagLabelModelType{Model type} indicates the type of language models used, while \tagLabelTechniques{Technique} describes the strategies adopted to address the task.  \tagLabelLoss{Training information} specifies the learning objectives applied during fine-tuning, \tagLabelLLM{LLM-related} highlights the specific use of large language models within the systems developed, and \tagLabelBias{Bias mitigation} denotes methods explicitly designed to address gender bias. Finally, \tagLabelExternalData{External data} indicates the external data used to solve the task.

% Most systems were modeled as retrieval-based architectures leveraging semantic similarity, with some teams incorporating reranking modules using LLMs or cross-encoders. While some approaches involved fine-tuning the embedding space, others did not. Embedding fine-tuning was performed using contrastive learning techniques, including losses such as GIST and InfoNCE, as well as more advanced strategies like curriculum learning. Large language models (LLMs) such as Gemma, Claude, and Llama 3.1 were utilized for both translation and reranking purposes. Regarding the use of external resources, teams such as pjmathematician employed ESCO data for Chinese translation, while NLPnorth and Talentrophy further enriched their systems with additional information from ESCO. With respect to bias mitigation, most teams did not implement explicit strategies. The exception was Ixa, which translated job titles into English to remove gender markers before performing reranking.

% Table of Approaches
\begin{table}[htbp]
\centering
\caption{Overview of team approaches for Task A .}
\resizebox{\textwidth}{!}{%
\label{tab:participant_task_a}
\begin{tabular}{@{}llp{13cm}@{}}
\toprule
\textbf{Team} & \textbf{Ref} & \textbf{Techniques} \\
\midrule
pjmathematician & \cite{vachharajani2025pjmathematician} &
\tagModelType{Multilingual Encoder-based Embedding Model}\tagTechniques{Retrieval}\tagLoss{LoRA}

\tagModelType{Multilingual Decoder-based Embedding Model}\tagLoss{Contrastive Learning}
\tagLoss{MSE Loss}
\tagLLM{Translation with LLMs}
\tagExternalData{ESCO Data for translation} \\
\midrule
NLPnorth & \cite{zhang2025nlpnorth} &
\tagModelType{Multilingual Encoder-based Embedding Model}\tagTechniques{Retrieval}\tagTechniques{Prompting}
\tagModelType{Multilingual Decoder-based Embedding Model}\tagTechniques{Classification}
\tagLoss{InfoNCE Loss}\tagLoss{Classification Loss}\tagExternalData{ESCO job descriptions} \\
\midrule
AlexU-NLP & \cite{barakat2025alexunlp} &
\tagModelType{Multilingual Encoder-based Embedding Model}\tagTechniques{Hybrid retrieval}
\tagTechniques{Reranking}\tagLoss{Curriculum Learning}\tagLoss{InfoNCE Loss}
\tagLLM{Reranking with LLM}
\tagExternalData{ESCO metadata} \\
\midrule
NT & \cite{ho2025ntteam} &
\tagModelType{Multilingual Encoder-based Embedding Model}\tagTechniques{Retrieval}\tagTechniques{Prompting}
\tagLoss{Zero-shot}\\
\midrule
DS@GT TalentCLEF & \cite{brikman2025multilingualjob} &
\tagModelType{Encoder-based Embedding Model}\tagTechniques{Retrieval}
\tagLoss{Contrastive Learning}\\
\midrule
Ixa & \cite{rodriguez2025twostage}{} &
\tagModelType{Encoder-based Embedding Model}\tagTechniques{Retrieval}\tagTechniques{Prompting}\tagLoss{Zero-shot}
\tagLLM{Reranking with LLM}\tagLLM{Translation with LLMs}\tagBias{Remove gender markers}\\
\midrule
SkillSeekers & \cite{uddin2025enhancing}  &  \tagModelType{Multilingual Encoder-based Embedding Model}\tagTechniques{Retrieval}
\tagLoss{Contrastive Learning}\\
\midrule

SCaLAR& \cite{bhangale2025finetuned} &  \tagModelType{Multilingual Encoder-based Embedding Model}\tagTechniques{Retrieval}
\tagLoss{Contrastive Learning}\tagLoss{InfoNCE Loss}\tagExternalData{ESCO job descriptions}\\
\midrule
VerbaNexAI & \cite{moreno2025verbanex} &  \tagModelType{Multilingual Encoder-based Embedding Model}\tagTechniques{Retrieval}
\tagLoss{Contrastive Learning}\\
\midrule
UDII-UPM & \cite{rodriguezvidal2025udiiupm} & \tagModelType{Encoder-based Embedding Model} \tagTechniques{Retrieval}\tagTechniques{Ensemble} \tagLoss{Zero-shot}  \\
\midrule
TechWolf &  \cite{decorte2025techwolf} &  \tagModelType{Multilingual Encoder-based Embedding Model}\tagTechniques{Retrieval}
\tagLoss{Contrastive Learning}\tagLoss{InfoNCE Loss}\tagLoss{Asymmetric dense projections}
\tagExternalData{Internal job ads}\tagLLM{Translation with LLMs}\\\midrule
HULAT-UC3M & \cite{tejera2025hulatuc3m} &  \tagModelType{Multilingual Encoder-based Embedding Model}\tagTechniques{Retrieval}\tagTechniques{Reranking}
\tagTechniques{Prompting}\tagLoss{Contrastive Learning}\tagLoss{InfoNCE Loss}\tagLoss{Zero-shot}
\tagLLM{Reranking with LLMs}\tagBias{Mitigation with LLM prompting}\\
%BK9999 &  &  \\\midrule
%Hallucinators &  &  \\\midrule
%ATV &  &  \\\midrule
%Reveries &  &  \\\midrule
%Edge &  &  \\\midrule
%grsilva &  &  \\\midrule
\bottomrule
\end{tabular}
}
\end{table}

Most of the participating teams approached the task as a retrieval problem, employing embedding models to compute semantic similarity as the ranking metric. Some teams, such as AlexU-NLP and Ixa, incorporated reranking modules based on LLMs, while AlexU-NLP also experimented with cross-encoders. Many teams fine-tuned the embeddings in the training data using contrastive learning techniques and loss functions such as GIST~\cite{solatorio2024gistembed} and InfoNCE~\cite{oord2018representation}, as well as more advanced strategies such as curriculum learning~\cite{soviany2022curriculum}, as employed in the AlexU-NLP submissions.

In general, encoder-based approaches predominated in the competition. However, there was notable experimentation with large-scale decoder-based embedding models. Rather than extensive fine-tuning of foundational models, most teams relied on state-of-the-art multilingual encoders with hundreds of millions of parameters, including architectures such as bge-m3~\cite{chen2024m3}, the multilingual-e5 family~\cite{wang2024multilingual}, and the GTE family~\cite{li2023towards}. Some participants, such as pjmathematician and NLPnorth, also incorporated large decoder-based models in some of their submissions, specifically gte-Qwen2-7B-instruct and Linq-Embed-Mistral~\cite{choi2024linq}. LLMs  (Gemma 2~\cite{team2024gemma}, Claude Sonnet 3.7, Qwen2.5~\cite{tahmid2024qwen2}, Llama 3.1~\cite{grattafiori2024llama}, and gpt-4.1-nano) were used  for machine translation by pjmathematician, Ixa, and TechWolf, and for reranking by Ixa, HULAT-UC3M, and AlexU-NLP.

Regarding external resources, participants were allowed to use additional information beyond the training data provided. Pjmathematician translated the ESCO data to adapt the embeddings for Chinese, while NLPnorth and AlexU-NLP enriched their data with additional information from the ESCO taxonomy. TechWolf opted to use its internal labor-market domain data to train their model. With respect to bias mitigation, most teams did not implement specific strategies; the exception was Ixa, which automatically translated Spanish and German job titles into English to remove gender markers.

% Modelización como sistema de Retrieval, usando similitud semántica, y en algunos casos con reranking usando LLMs o crossencoders. 
% ALgunos ajustan embedding y otros no. 
%Ajuste del embedding utilizando contrastive learning techniques incluyendo losses como GIST, INFONCE, y tecnicas más avanzadas de apredizaje comoCurriculum Learning. 
% LLMs para traducción y para reranking.  Usando Gemma, Claude y Llama3.1. 
% En relación a dots externos, utilian pjmathematician ESCO Data para traducir al chino,, NLPnortj y Talentrphy utilizan más información de ESCO para enriquecer datos.
% En relación a la gestión del bias, la mayoría deequipos no hacen nada especifco a ecepción d eIxa, que traduce al inglés para eliminar esos markadores de generos antes de hacer le reranking. 

\subsection{Task B}
Task B had 68 registered teams, with 10 teams submitting at least one run during the evaluation phase, and 8 being part of the final benchmark, for a total of 84 Codabench submissions. Table~\ref{tab:participant_task_b} summarizes the methods used by the participating teams, following the same color-code as in the previous section. 

%La tarea B contó con 68 equipos registrados, con 10 equipos subiendo al menos un run en la fase de evalaución, y contribuyendo con un total de 84 submissions. La Table~\ref{tab:participant_task_b} muestra los métodos utilizados por los equipos participantes con la misma representación gráfica de colores que se utilizó en la sección anterior. 

\begin{table}[htb]
\centering
\caption{Overview of team approaches for Task B with keywords labeled by color-coded category.}
\resizebox{\textwidth}{!}{%
\label{tab:participant_task_b}
\begin{tabular}{@{}llp{13cm}@{}}
\toprule
\textbf{Team} & \textbf{Ref} & \textbf{Techniques} \\
\midrule
pjmathematician & \cite{vachharajani2025pjmathematician} & 
\tagModelType{Multilingual Encoder-based Embedding Model}\tagTechniques{Retrieval}\tagTechniques{Prompting}
\tagLoss{Contrastive Learning} \tagLoss{GIST Loss}\tagLoss{LoRA}
\tagLLM{Data Augmentation using LLMs} \\
\midrule
NLPnorth & \cite{zhang2025nlpnorth} & 
\tagModelType{Multilingual Encoder-based Embedding Model}\tagTechniques{Retrieval}\tagTechniques{Prompting}
\tagModelType{Multilingual Decoder-based Embedding Model}\tagTechniques{Classification}
\tagLoss{Contrastive Learning}\tagLoss{InfoNCE Loss}\tagLoss{Classification Loss}
\tagExternalData{ESCO skill and jobs descriptions}\\
\midrule
iagox & \cite{vazquezgarcia2025beyondtitles} & \tagModelType{Encoder-based Embedding Model}\tagTechniques{Retrieval}\tagTechniques{Ensemble}\tagLoss{Zero-shot}
\tagLLM{Definitions generation using LLMs}
\\
\midrule
moali & \cite{ali2025enhancingjob} & \tagModelType{Multilingual Encoder-based Embedding Model}\tagTechniques{Retrieval}
\tagLoss{Contrastive Learning}\tagLoss{InfoNCE}\tagLLM{Definition generation using LLMs}
\tagExternalData{ESCO skill descriptions}\\\midrule
SkillSeekers & \cite{uddin2025enhancing}  &  \tagModelType{Multilingual Encoder-based Embedding Model}\tagTechniques{Retrieval}\tagLoss{Zero-shot}
\\\midrule
TechWolf &  \cite{decorte2025techwolf} &  \tagModelType{Multilingual Encoder-based Embedding Model}\tagTechniques{Retrieval}
\tagLoss{Contrastive Learning}\tagLoss{InfoNCE Loss}\tagLoss{Asymmetric dense projections}
\tagExternalData{Internal job ads}\tagLLM{Translation with LLMs}\\\midrule
HULAT-UC3M & \cite{tejera2025hulatuc3m} &  \tagModelType{Multilingual Encoder-based Embedding Model}\tagTechniques{Retrieval}\tagTechniques{Reranking}
\tagTechniques{Prompting}\tagLoss{Zero-shot}
\tagLLM{Reranking with LLMs}\\\midrule
COTECMAR-UTB & \cite{llamas2025cotecmarutb} &  \tagModelType{Multilingual Encoder-based Embedding Model}\tagTechniques{Retrieval}\tagLoss{Zero-shot}
\\
%Hallucinators & Ref  &  \\\midrule
%Reveries & Ref &  \\\midrule

\bottomrule
\end{tabular}
}
\end{table}

Given the similarity in the input data, most participants approached Task B using methodologies very similar to those employed in Task A. The task has been solved as a semantic similarity-based retrieval problem, trying to identify the most relevant skills for a given job title. Several teams~\cite{vachharajani2025pjmathematician,zhang2025nlpnorth}, used encoder models similar to those used in Task A, but fine-tuned with the provided training data by means of contrastive learning techniques. Additionally, prompting strategies were integrated into the pipelines to obtain better contextualized embeddings for retrieval purposes. 

LLMs were commonly used to enrich the textual representation of skills by generating synthetic definitions~\cite{ali2025enhancingjob,vazquezgarcia2025beyondtitles}, with the goal of enhancing semantic coverage and improving retrieval performance.  For data augmentation, participants mainly used models from the Qwen2.5~\cite{tahmid2024qwen2} and Llama 3.1~\cite{grattafiori2024llama} families. External resources like descriptions from the ESCO taxonomy were leveraged by teams such as NLPnorth and moali to augment the training data. As in Task A, most of the systems relied on multilingual encoder-based models, although the task is monolingual.  In some cases, these models were used in zero-shot settings~\cite{uddin2025enhancing,vazquezgarcia2025beyondtitles}.

% https://geographyfieldwork.com/RadarChartCreator.html
%\begin{figure}[h]
%\caption{Distribution of used features}
%\label{fig:distribution_participants}
%\includegraphics[width=\textwidth]{plot}
%\end{figure}

% https://app.datawrapper.de/edit/uCSzk/basemap
%\begin{figure}[h]
%\caption{Distribution of used features}
%\label{fig:participants}
%\includegraphics[width=\textwidth]{participants}
%\end{figure}

\section{Results}
% Maybe introduce the section with general metrics and comunication
% campaign done? (In case we want to use space). Maybe a map of 
% the countries of the participating teams.
% 
% A continuación se presentan los resultados obtenidos para las tareas A y B. Debido al elevado número de sistemas recibidos, en las tablas se reportan únicamente las mejores submissions de cada equipo cuyo rendimiento superó al baseline. Los resultados completos, incluyendo todas las submissions de cada equipo, están disponibles en la página web oficial de la tarea.

 The results obtained for tasks A and B are presented below. Given the large number of systems received, only the best submissions from each team are reported in the tables. The complete results, including all submissions from each team for tasks A and B, are available on the official task web page\footnote{Complete results for Task A: 
\url{https://talentclef.github.io/talentclef/docs/talentclef-2025/results/task\_a\_results/}
 
Complete results for Task B: \url{https://talentclef.github.io/talentclef/docs/talentclef-2025/results/task\_b\_results/}}%
 
In Task A, teams were evaluated in four areas: overall multilingual performance, cross-lingual performance, Chinese language results (which were optional due to lack of training data), and gender bias. With this evaluation setup, we could analyze not just performance, but also how well each system handled multilingual challenges, cross-lingual transfer, and fairness issues. In contrast, Task B focused exclusively on the evaluation of systems in English.

\subsection{Task A Main Results}
% Partiicpants, systems and results
% Show table of top-performing teams (Average en-de-es). Include best system per team, only top-10. 

%  En la tarea A, se ha llevado una evaluación de los equipos según su rendimiento multilingue general, su rendimiento cross-lingue,sistemas que decidieron ajustar para chino (opcional, por no disponer de datos de entrenamiento)  y el resultado a nivel de sesgo. 

% En la tarea A los participantes podían enviar los resultados de hasta 5 sistemas diferentes. Para la generación de los benchmarks hemos utilizado el sistema de mejor calificación de cada uno de los equipos que ha superado el baseline, y llevar los naálisis. Por una parte, se buscaba evaluar el sistema con mejor ufncionamieno en varios idiomas incluyendo inglés, español y alemán. Los resultados de esto pueden mostrarse en Table X, donde se observa que el top performer team es talenttropy, con Techwolf como segundo mejor sistema overall.
\subsubsection{Overall Multilingual Performance.}

The main leaderboard, shown in Table~\ref{tab:team_results_task_a}, reports the Mean Average Precision across monolingual scenarios in English, Spanish, and German. \textbf{AlexU-NLP} achieved the best overall multilingual performance, with an average MAP of 0.534 and the best results for Spanish and German. \textbf{TechWolf} was the second-best team overall (0.517), demonstrating balanced and strong results in the three languages considered. It is notable that both \textbf{pjmathematician} and \textbf{NLPnorth} achieved strong results in English, with NLPnorth reaching a MAP of 0.537 and pjmathematician obtaining the highest score in that language (0.563).

%Los sistemas con mejor rendimiento comparten como característica principal el ajuste del espacio de embeddings utilizando los datos de entrenamiento, así como el uso de Losses que han demostrado una alta eficacia en tareas de similitud semántica, como GIST o InfoNCE. A Aunque el equipo Talentrphy empleó técnicas de retrieval híbrido y reranking en algunas de sus submissions, en esta en particular no se reportó el uso de sistemas de reranking. No obstante, sí evitaron explícitamente el uso de muestras crosslingües durante el entrenamiento
% Por otra parte, el equipo Techwolf desarrolló un sistema contrastivo que utilizaba proyecciones densas asimétricas para pares formados por job titles y conjuntos de skills separados por comas. Este enfoque, utilizando datos propios, le permitió opbtener buenos resultados, alcanzando la segunda mejor posición en la clasificación general (overall) y también la segunda mejor posición en los subtasks en español y alemán.

The top-performing systems shared the common feature of fine-tuning their embedding spaces using training data, and leveraging loss functions that have demonstrated strong effectiveness in semantic similarity tasks, such as GIST or InfoNCE. Although the AlexU-NLP team explored hybrid retrieval and re-ranking techniques in some of their submissions, re-ranking was not used in the submission reported here. However, they explicitly avoided incorporating cross-lingual samples during training. On the other hand, the TechWolf team developed a contrastive approach that employs dense asymmetric projections for pairs consisting of job titles and sets of skills separated by commas. This strategy, which relies on their internal and proprietary data, allowed them to achieve strong results, securing the second position in the overall ranking, as well as the second best performance in both the Spanish and German languages.

\begin{table}[htb!]
\centering

\caption{Overview of team results for Task A. Best value per column is in bold, second best  is underlined.}
\resizebox{1\textwidth}{!}{%
\label{tab:team_results_task_a}
\begin{tabular}{@{}l @{\hspace{0.2cm}} 
                c @{\hspace{0.2cm}}
                c @{\hspace{0.2cm}}
                c @{\hspace{0.2cm}}
                c @{\hspace{0.2cm}}
                c@{}}
\toprule
\textbf{Team} & \textbf{System ID} & \textbf{Avg} & \textbf{MAP(en-en)} & \textbf{MAP(es-es)} & \textbf{MAP(de-de)} \\
\midrule
\rowcolor{gold!20} AlexU-NLP           & 283782  & \textbf{0.534} & \underline{0.559} & \textbf{0.527} & \textbf{0.516} \\
\rowcolor{silver!20} TechWolf           & 284991  & \underline{0.517} & 0.533 & \underline{0.519} & \underline{0.500} \\
pjmathematician         & 275330  & 0.515 & \textbf{0.563} & 0.507 & 0.476 \\
NLPnorth                & 276154  & 0.492 & 0.537 & 0.496 & 0.442 \\
NT                      & 284052  & 0.464 & 0.523 & 0.466 & 0.404 \\
%BK9999                  & 273291  & 0.460 & 0.477 & 0.439 & 0.464 \\
%Reveries                & 283752  & 0.449 & 0.463 & 0.438 & 0.445 \\
SCaLAR                  & 284749  & 0.446 & 0.473 & 0.438 & 0.427 \\
%Hallucinators           & 278926  & 0.444 & 0.469 & 0.445 & 0.417 \\
HULAT-UC3M              & 279708  & 0.420 & 0.479 & 0.420 & 0.360 \\
UDII-UPM                & 276052  & 0.414 & 0.448 & 0.415 & 0.377 \\
DS@GT - TalentCLEF      & 285019  & 0.399 & 0.440 & 0.402 & 0.355 \\
%Edge                    & 284303  & 0.366 & 0.354 & 0.372 & 0.373 \\
VerbaNexAI              & 284612  & 0.360 & 0.408 & 0.348 & 0.324 \\
TalentCLEF Baseline     & ---     & 0.360 & 0.408 & 0.348 & 0.324 \\
SkillSeekers     & 284999     & 0.354 & 0.355 & 0.360 & 0.347 \\
Ixa     & 285265     & 0.181 & 0.199 & 0.173 & 0.169 \\
\bottomrule
\end{tabular}%
}
\end{table}

\subsubsection{Cross-Lingual Performance.}

% Para evaluar el rendimiento en escenarios cross-lingues, se consideró como métrica el MAP promedio entre los escenarios con queries en ingles y corpus elements en español y aleman. . Los resultados se presentan en la Tabla~\ref{tab:team_results_task_a_cross}. Las submissions de los equipos \textbf{pjmathematician} y \textbf{Talentropy} fueron las más destacadas, obteniendo la misma puntuación final, aunque cada uno alcanzó el mejor resultado en uno de los dos idiomas considerados..

%Como era de esperar,  los valores de MAP en los escenarios cross-linguales fueron, en general, inferiores a los obtenidos en el caso monolingüe. Sin embargo, los resultados siguen siendo relativamente altos, lo que refleja un progreso significativo en la similitud semántica multilingüe, aunque también confirma que la transferencia semántica entre lenguas continúa siendo un reto para los sistemas actuales. Cabe destacar que el equipo \textbf{pjmathematician}, que obtuvo la puntuación más alta, empleó un modelo de embeddings basado en la arquitectura GTE, ajustado a partir de un modelo tipo decoder de 7 mil millones de parámetros. En contraste, \textbf{Talentropy} utilizó en esta submission una estrategia de recuperación híbrida, combinando BM25 con modelos de embeddings basados en E5, y aplicando un reranking mediante un cross-encoder afinado específicamente para esta tarea. Ambos equipos obtuvieron la misma puntuación final, lo que sugiere que el uso de modelos de gran tamaño no implica necesariamente una mejora en el rendimiento para este tipo de tareas.

To evaluate performance in cross-lingual settings, the metric considered was the average MAP for the two scenarios of English queries vs. corpus elements in Spanish and German, respectively. The results are presented in Table~\ref{tab:team_results_task_a_cross}. The submissions from the \textbf{pjmathematician} and \textbf{AlexU-NLP} teams achieved the highest scores, with both teams reaching the same final score (0.514), although each obtained the best result in each of the two language pairs.

As expected, MAP values in cross-lingual scenarios were generally lower than in the monolingual case. However, the results are relatively high, showing important progress in multilingual semantic similarity but also confirming that semantic transfer between languages is still a challenge for today's systems. It should be noted that the pjmathematician team, which achieved the highest score, used an embedding model based on the GTE architecture fine-tuned from a 7 billion-parameter decoder model. In contrast, AlexU-NLP employed a hybrid retrieval strategy in this submission, combining BM25 with E5-based embedding models and applying re-ranking with a cross-encoder specifically fine-tuned for this task. The final results were similar, showing that the use of a larger model does not necessarily imply better results in this type of task.

\clearpage
\begin{table}[htb]
\centering

\caption{Team results for (en-es) and (en-de) cross-lingual track. Best value per column in bold, second in underline.}
\label{tab:team_results_task_a_cross}
\begin{tabular}{@{}l @{\hspace{0.2cm}}
                c @{\hspace{0.2cm}}
                c @{\hspace{0.2cm}}
                c @{\hspace{0.2cm}}
                c @{}}
\toprule
\textbf{Team} & \textbf{System ID} & 
\textbf{Avg} & \textbf{MAP(en-es)} & \textbf{MAP(en-de)} \\
\midrule
\rowcolor{gold!20} pjmathematician & 275330 & \textbf{0.514} & \textbf{0.525} & \underline{0.504} \\
\rowcolor{silver!20} AlexU-NLP & 284479 & \underline{0.514} & \underline{0.516} & \textbf{0.512} \\
TechWolf & 284991 & 0.504 & 0.510 & 0.498 \\
NLPnorth & 276154 & 0.477 & 0.492 & 0.461 \\
%Reveries & 283752 & 0.434 & 0.433 & 0.435 \\
%BK9999 & 273291 & 0.417 & 0.395 & 0.439 \\
%Hallucinators & 278926 & 0.395 & 0.382 & 0.407 \\
HULAT-UC3M & 279708 & 0.391 & 0.416 & 0.365 \\
UDII-UPM & 276052 & 0.379 & 0.383 & 0.375 \\
SCaLAR & 284749 & 0.368 & 0.366 & 0.370 \\
%Edge & 284303 & 0.354 & 0.354 & 0.353 \\\midrule
TalentCLEF Baseline & --- & 0.340 & 0.335 & 0.345 \\
VerbaNexAI & 284612 & 0.339 & 0.335 & 0.344 \\
% TalentCLEF Baseline & --- & 0.340 & 0.335 & 0.345 \\

\bottomrule
\end{tabular}%
\end{table}

\subsubsection{Systems including Chinese language}

%Debido a la ausencia de datos de entrenamiento para chino, el envío de resultados para este idioma fue de carácter voluntario en la tarea. A pesar de ello, la mayoría de los equipos participantes (13) renviaron predicciones para esta lengua, reportando varios de ellos el uso de LLMs para llevar a cabo la traducción automática del inglés para obtener datos de entrenamiento, como se describió en la sección 2.

Due to the absence of training data for Chinese, submission of results for this language was voluntary in the task. Despite this, most of the participating teams (9) submitted predictions for this language, and several of them reported the use of LLMs to do machine translation from English to obtain training data, as described in Section \ref{subsec:participants_taskA}.

%La Tabla~\ref{tab:team_results_task_a_zh} muestra los resultados de los sistemas recibidos, ordenados por el MAP promedio (Avg. MAP). En este escenario, el equipo \textbf{TalentTrophy} obtuvo el mejor MAP medio, seguido por \textbf{Techwolf} y \textbf{pjmathematician}, ambos en la misma posición global. El rendimiento general de TalentTrophy se ve favorecido por sus buenos resultados en español y alemán, ya que su desempeño en chino fue inferior en en comparación a esos idiomas. Por otra parte, Techwolf consiguió tanto el segundo mejor resultado promedio como el segundo mejor resultado específico en chino. Finalmente, \textbf{pjmathematician} logró el mejor rendimiento en chino ,al igual que en inglés, aunque su promedio global se vio penalizado por un rendimiento más bajo en alemán y español.

Table~\ref{tab:team_results_task_a_zh} shows the results for all systems submitted, sorted by average MAP now including Chinese. In this scenario, \textbf{AlexU-NLP} achieved the highest score with 0.527, followed by \textbf{TechWolf} and \textbf{pjmathematician},  sharing the second position with 0.515. The final result for AlexU-NLP is influenced by its stronger performance in Spanish and German, while its score in Chinese is lower than in the other languages. TechWolf obtained the second-best result both in terms of average and specifically for Chinese. Pjmathematician achieved the best score in Chinese, as well as in English, but did not reach a higher overall ranking due to lower performance in Spanish and German.

\begin{table}[htb]
\centering
\caption{Overview of team results for Task A submissions including Chinese. Best value per column in bold, second in underline.}
\resizebox{1\textwidth}{!}{%
\label{tab:team_results_task_a_zh}
\begin{tabular}{@{}l @{\hspace{0.2cm}}
                c @{\hspace{0.2cm}}
                c @{\hspace{0.2cm}}
                c @{\hspace{0.2cm}}
                c @{\hspace{0.2cm}}
                c @{\hspace{0.2cm}}
                c @{\hspace{0.2cm}}
                c @{}}
\toprule
\textbf{Team} & 
\textbf{System ID} & 
\textbf{Avg} & 
\textbf{MAP(en-en)} & 
\textbf{MAP(es-es)} & 
\textbf{MAP(de-de)} & 
\textbf{MAP(zh-zh)} \\
\midrule
\rowcolor{gold!20} AlexU-NLP          & 283782  & \textbf{0.527}   & \underline{0.559}  & \textbf{0.527}   & \textbf{0.516}   & 0.508 \\
\rowcolor{silver!20} pjmathematician        & 275330  & \underline{0.515}   & \textbf{0.563}  & 0.507          & 0.476          & \textbf{0.516} \\
\rowcolor{silver!20} TechWolf          & 284991  & \underline{0.515}    & 0.533          & \underline{0.519}    & \underline{0.500}    & \underline{0.510} \\
NLPnorth               & 276154  & 0.492   & 0.537           & 0.496          & 0.442          & 0.495  \\
NT                     & 284052  & 0.473   & 0.523           & 0.466          & 0.404          & 0.497 \\
%BK9999                 & 273291  & 0.463   & 0.477           & 0.439          & 0.464          & 0.473 \\
%Hallucinators          & 278926  & 0.450   & 0.469           & 0.445          & 0.417          & 0.469 \\
%Reveries               & 283752  & 0.450   & 0.463           & 0.438          & 0.445          & 0.451 \\
%Reveries               & 284547  & 0.427   & 0.443           & 0.413          & 0.418          & 0.436 \\
UDII-UPM               & 276052  & 0.423   & 0.448           & 0.415          & 0.377          & 0.451 \\
HULAT-UC3M             & 280184  & 0.418   & 0.479           & 0.420          & 0.360          & 0.415 \\
DS@GT - TalentCLEF     & 285019  & 0.405   & 0.440           & 0.402          & 0.355          & 0.421 \\
SkillSeekers           & 284999  & 0.368   & 0.355           & 0.360          & 0.347          & 0.412 \\
%Edge                   & 284303  & 0.366   & 0.354           & 0.372          & 0.373          & 0.367 \\
\bottomrule
\end{tabular}%
}
\end{table}

\subsubsection{Bias Evaluation}

In addition to standard retrieval metrics, we performed a bias analysis by measuring performance gaps between gendered job titles. Systems that minimize this disparity are recognized as biased-controlled and as promoting fairness alongside performance. To rank the systems, we used the average MAP across German and Spanish job titles, the two languages in which grammatical gender can introduce bias. Furthermore, we evaluated the consistency of ranking for gendered job titles using the RBO metric between masculine and feminine forms in both Spanish and German. The average values are shown in Table ~\ref{tab:bias_results_es_de}. 

In this setting, both \textbf{NLPnorth} and \textbf{Ixa} achieved a perfect score in both Spanish and German, which means that their system outputs were identical regardless of gendered lexical changes in the input. In the case of Ixa, all job titles were translated into English, thus eliminating grammatical gender marks, given that English is a \textit{natural gender} language\footnote{Languages in which gender is made explicit mostly through pronouns (e.g. \textit{hers} versus \textit{his}) or in some lexical cases (e.g. \textit{queen} versus \textit{king}).}. On the other hand, despite NLPnorth not reporting any specific strategy for gender bias mitigation, their classification-based system produced fully stable rankings while also maintaining high MAP scores in both languages. Overall, the results show that the best performing systems (AlexU-NLP and TechWolf) are highly effective in minimizing gender bias in their output classifications, with average RBO values above 0.97, even in the absence of explicit bias mitigation techniques.

%\textbf{Talentropy}, \textbf{TechWolf}, and \textbf{BK9999} also demonstrated high overall performance and relatively high RBO values (above 0.95 on average). However, their minimum RBO scores occasionally dipped below 0.90 in at least one language, suggesting some minor sensitivity to gender-specific variations, despite their overall robustness.

%\textbf{pjmathematician}, \textbf{Reveries}, \textbf{SCaLAR}, and \textbf{Hallucinators} displayed moderate consistency, with average RBO values ranging from approximately 0.94 to 0.97. However, their minimum RBO scores show more pronounced variability, in some cases dropping below 0.80, indicating increased instability in particular cases or job titles.

%Other teams, such as \textbf{NT}, \textbf{HULAT-UC3M}, \textbf{VerbaNexAI}, and \textbf{Edge}, showed greater variability and lower minimum RBO scores—falling below 0.70 in some instances—reflecting inconsistent treatment of gendered forms and a higher degree of gender bias sensitivity.

%Overall, these results suggest that while several systems achieve high effectiveness with minimal gender bias, a subset of participants still experience notable ranking instability for gendered job titles, underscoring the importance of addressing lexical gender variation in multilingual information retrieval.

% : Average MAP and Rank-Biased Overlap (RBO) by team and run\_ID. \textbf{AVG MAP (es-de)} is the average MAP for es-es and de-de; \textbf{AVG RBO} is the mean of rbo\_avg (es) and rbo\_avg (de). Best value per metric in bold, second-best in underline.
\begin{table}[htb]
\centering
\caption{Performance considering fairness. Best value per column are in bold, second best are underlined.}
\label{tab:bias_results_es_de}
\resizebox{1\textwidth}{!}{%
\begin{tabular}{@{}l c c c c c@{}}
\toprule
% \label{tab:bias_results_es_de}
\textbf{Team} & \textbf{System ID} & \textbf{Avg. MAP(es \& de)} & \textbf{Avg. RBO (es-de)} & \textbf{RBO(es)} & \textbf{RBO(de)} \\
\midrule
\rowcolor{gold!20} AlexU-NLP         & 283782  & \textbf{0.521}  & \underline{0.976}  & \underline{0.975} & \underline{0.977} \\
\rowcolor{silver!20} TechWolf         & 284991  & \underline{0.510}   & 0.971          & 0.972         & 0.970 \\
pjmathematician       & 275330  & 0.492           & 0.947          & 0.949         & 0.945 \\
NLPnorth              & 276154  & 0.469           & \textbf{1.000} & \textbf{1.000} & \textbf{1.000} \\
%BK9999                & 273291  & 0.451           & 0.968          & 0.969         & 0.967 \\
%Reveries              & 283752  & 0.442           & 0.973          & 0.974         & 0.973 \\
NT                    & 284052  & 0.435           & 0.780          & 0.818         & 0.742 \\
SCaLAR                & 284749  & 0.433           & 0.947          & 0.956         & 0.938 \\
%Hallucinators         & 278926  & 0.431           & 0.949          & 0.959         & 0.939 \\
UDII-UPM              & 276052  & 0.396           & 0.932          & 0.925         & 0.939 \\
HULAT-UC3M            & 279708  & 0.390           & 0.948          & 0.950         & 0.946 \\
DS@GT - TalentCLEF    & 285019  & 0.379           & 0.934          & 0.938         & 0.929 \\
%Edge                  & 284303  & 0.372           & \textbf{0.984} & \textbf{0.995} & 0.974 \\
SkillSeekers          & 284999  & 0.354           & 0.966          & 0.974         & 0.957 \\
VerbaNexAI            & 284612  & 0.336           & 0.915          & 0.893         & 0.937 \\
TalentCLEF Baseline   & 281638  & 0.336           & 0.915          & 0.893         & 0.937 \\
Ixa                   & 285265  & 0.171           & \textbf{1.000} & \textbf{1.000} & \textbf{1.000} \\
\bottomrule
\end{tabular}%
}
\end{table}

\subsection{Task B Main Results}
% Partiicpants, systems and results

%La Tabla~\ref{tab:task_b_final_map} muestra los resultados de los equipos participantes según los valores de MAP. pjmathematician obtuvo el MAP más alto (0.360), seguido de TeamB (0.345), con el resto de quipos en un nivel de rendimiento . 
Table~\ref{tab:task_b_final_map} shows the MAP results for all participating teams in Task B. \textbf{Pjmathematician} achieved the highest MAP (0.360), followed by \textbf{moali} (0.345), with the remaining teams grouped at a lower performance level.

%Los top-performer teams utilizaron approaches similares a los de la Task A. En el caso de Pjmathematictician  un approach de retrieval con un embeddingde tipo decoder (7B), ajustado con una loss GIST utilizando tanto los datos de entenamiento como una aumentación de estos obtenidos mediante el modelo Qwen2.5 32B. El equipo TeamB ajustó un modelo Mulitlingual e5 con los datos de entrenamiento e incorporando las descripciones de skills de ESCO y generando datos de job descriptions obtenidos mediante prompting LLama3.1 8B. 

The top-performing teams in Task B followed strategies similar to those used in Task A. For pjmathematician, the approach was based on retrieval with a decoder-type embedding model (7B), fine-tuned using the GIST loss and leveraging both the provided training data and augmented data generated with the Qwen2.5 32B model. Moali fine-tuned a multilingual E5 model with training data, incorporating ESCO skill descriptions, and generating additional job descriptions via prompting with Llama 3.1 8B.

%El resto de los equipos optaron por métodos zero-shot, combinando modelos tipo encoder con técnicas de reranking basadas en modelos de lenguaje como Gemma3-4B y Gemini2.5Flash. En particular, Techwolf empleó el mismo modelo que le permitió obtener excelentes resultados en la Task A, logrando un buen rendimiento en esta tarea, especialmente considerando que no ajustó el modelo con datos de entrenamiento específicos.

The other teams mostly opted for zero-shot methods, combining encoder-based models with re-ranking techniques built on prompts using language models like Gemma3-4B and Gemini2.5Flash. In particular, Techwolf employed the same model architecture that led to strong results in Task A, achieving solid performance here as well, especially considering that their approach did not include additional fine-tuning with task-specific training data.

\begin{table}[htb]
\centering
\caption{Overview of team results for Task B submissions. Best value per column are in bold, second are underlined.}
\label{tab:task_b_final_map}
\resizebox{0.45\textwidth}{!}{%
\begin{tabular}{@{}l @{\hspace{0.2cm}} c @{\hspace{0.2cm}} c @{}}
\toprule
\textbf{Team} & \textbf{System ID} & \textbf{MAP} \\
\midrule
\rowcolor{gold!20} pjmathematician     & 278954  & \textbf{0.360} \\
\rowcolor{silver!20} moali               & 284333  & \underline{0.345} \\
NLPnorth            & 279245  & 0.290 \\
iagox               & 284949  & 0.278 \\
Techwolf            & 284828  & 0.265 \\
%Hallucinators       & 284969  & 0.253 \\
%Reveries            & 283514  & 0.233 \\
SkillSeekers        & 284784  & 0.224 \\
COTECMAR-UTB      & 284555  & 0.215 \\
%smontero            & 284300  & 0.205 \\
TalentCLEF Baseline        & 281657  & 0.196 \\
HULAT-UC3M.         & 280606  & 0.141 \\
\bottomrule
\end{tabular}
}
\end{table}

\subsection{Analytical Insights}
% Comparison between ewsults and size (again on models over the the baseline). 
% Maybe even an analysis of external data. 

% Tanto en la Tarea A como en la Tarea B se han realizado numerosas submissions generadas con diferentes sistemas y configuraciones. Como se muestra en la Tabla X y la Tabla Y, se han implementado sistemas que varían tanto en los parámetros empleados como en su modelización.
In both Task A and Task B, participants submitted a considerable number of systems with diverse configurations, as summarized in Table~\ref{tab:participant_task_a} and Table~\ref{tab:participant_task_b}. These systems cover a variety of techniques for training or using LLMs, reflecting the variety of approaches explored throughout the challenge. 

% En la Figura~\ref{fig:model_size_vs_map_TaskA.png} se presenta una gráfica que relaciona el tamaño de los modelos utilizados con su rendimiento medio en inglés, español y alemán, identificando tanto la estrategia de entrenamiento como la metodología empleada (Retrieval-only o Retrieval+Reranking). Además, se incluye una línea de regresión que permite visualizar la tendencia general.

\begin{figure}[ht!]
    \centering
    % \label{fig:model_size_vs_map_TaskA.png}
    \includegraphics[width=1\textwidth]{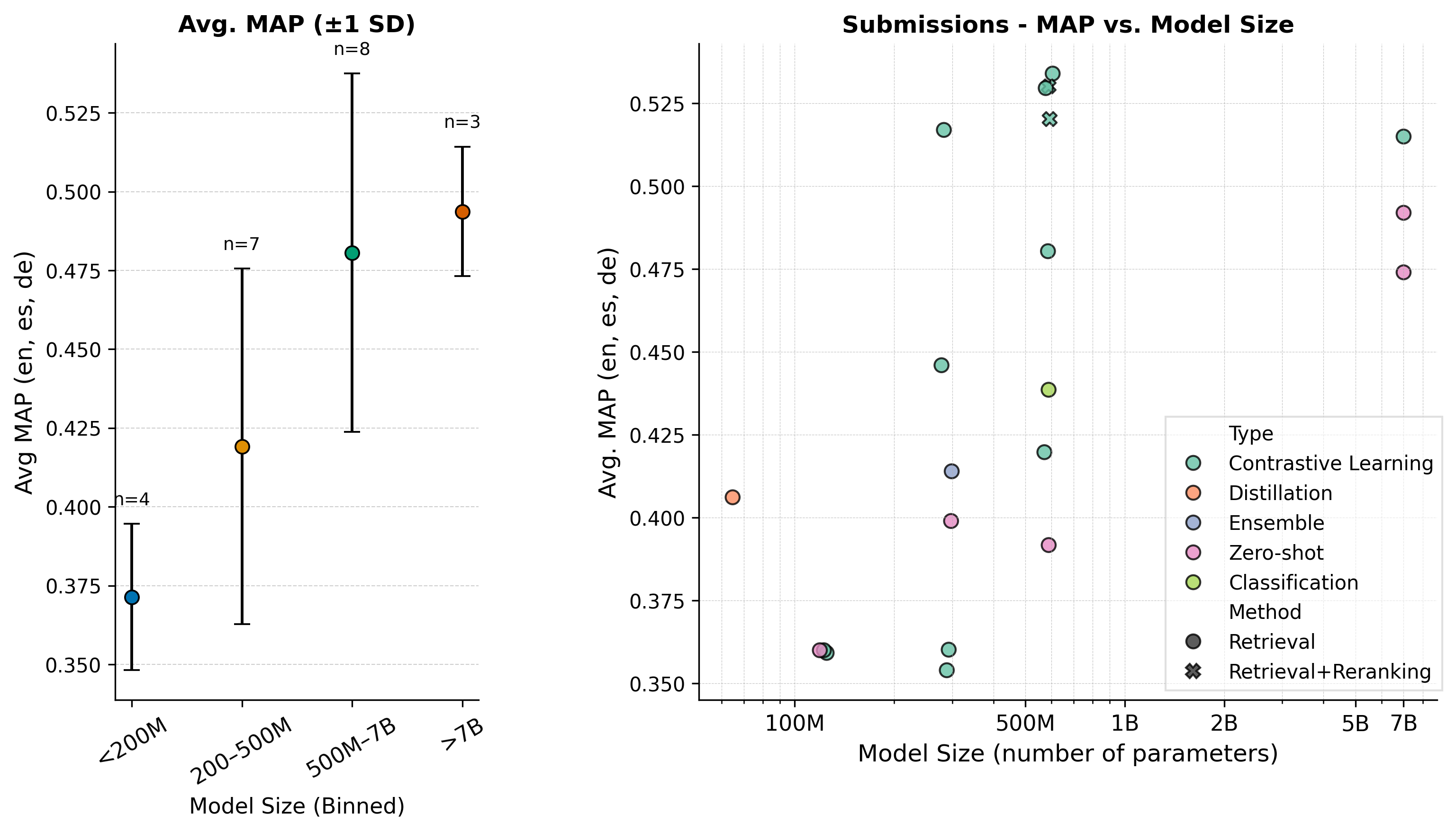}
    \caption{Relationship between model size and average Mean Average Precision (MAP) across English, Spanish, and German in Task~A. Left: Average MAP for submissions grouped by model size. Right: Individual submission results}
    \label{fig:model_size_vs_map}
\end{figure}

Figure~\ref{fig:model_size_vs_map} shows the relationship between the base model size (number of parameters) and the average MAP of system performance in English, Spanish, and German for Task~A. The plot on the left groups the systems into parameter-size bins and shows the average MAP with one standard deviation. The plot on the right shows individual system submissions, colored by training strategy and shaped by retrieval methodology. 

As seen in the left panel, we observe that, on average, there is a positive correlation between model size and system performance. However, despite this tendency, it should be noted that the highest MAP scores are obtained with models of around 500M parameters, which often outperform much larger decoder-based embedding models of up to 7B parameters. These results are particularly relevant when encoder-based models are fine-tuned using contrastive techniques and enhanced with a reranking approach. This suggests that model size alone does not guarantee optimal performance and that certain methodological choices, such as training strategy, play a fundamental role in system performance, even in models of smaller size.

% Now with Task B

%Por otra parte, la Figure X muestra la relación entre el tamaño delmodelo,  la estrategia de entreanmiento, y el uso de LLMs en la tarea, en relación al rendimiento obtenido en terminos de MAP
\begin{figure}[ht!]
    \centering
    \includegraphics[width=0.85\textwidth]{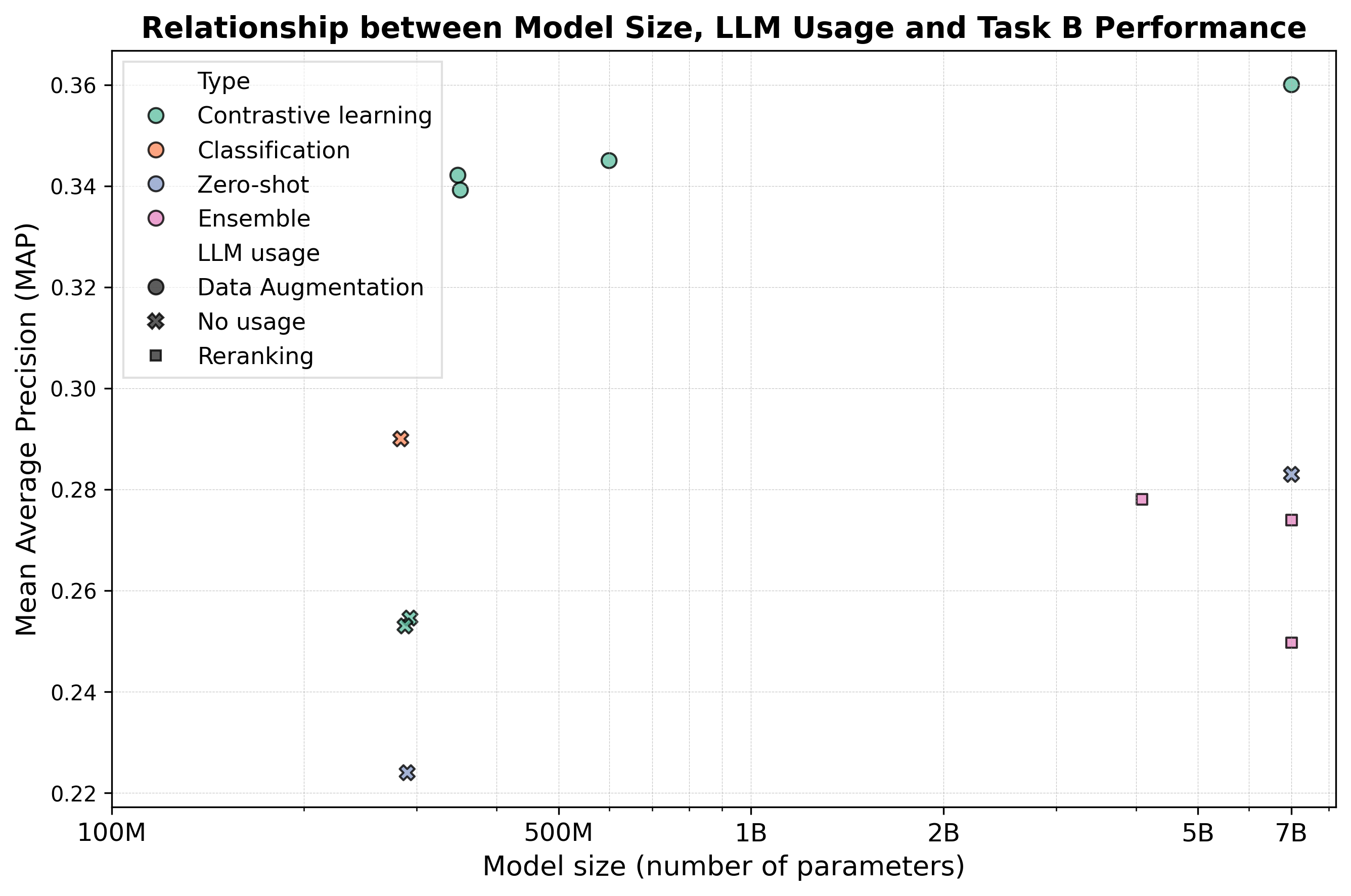}
    \caption{Relationship between model size (number of parameters) and average Mean Average Precision (MAP) in Task B.}
    \label{fig:model_size_vs_map_TaskB}
\end{figure}

On the other hand, Figure~\ref{fig:model_size_vs_map_TaskB}, shows the relationship between model size, training strategy, and the use of LLMs in the task, in relation to the resulting performance measured in terms of Mean Average Precision.

The image shows some interesting patterns. On the one hand, the use of LLMs for data augmentation, combined with fine-tuning through contrastive techniques, enables systems to achieve significantly higher performance, obtaining MAP improvements of up to 8 points for models of similar size. The results highlight the added value of using synthetic data generation with LLM together with specialized fine-tuning strategies.

%El análisis de la Figura~\ref{fig:model_size_vs_map_TaskB} revela varios patrones interesantes. Por un lado, se aprecia que el uso de LLMs para data augmentation, combinado con ajustes mediante técnicas contrastivas, permite obtener sistemas significativamente más performantes, logrando mejoras de hasta 8 puntos de MAP en modelos de tamaño similar. Este resultado evidencia el valor añadido de aprovechar la generación de datos sintéticos utilizando LLMs junto con estrategias de fine-tuning especializadas.

% Además, en este contexto específico—donde la relación entre job title y skills es más compleja que en la tarea de matching, se observa que los modelos de embedding de tipo decoder muestran una ligera mejora respecto a los encoders tradicionales, siempre que se acompañen de una adecuada estrategia de entrenamiento. Esta tendencia sugiere que, para tareas de detección de skill relevantes, los modelos de embedding construidos con LLMs y técnicas avanzadas de augmentación, pueden ofrecer ventajas adicionales frente a métodos más sencillos.

Moreover, in this specific context, where the relationship between job titles and skills is more complex than in Task A, decoder-based embeddings models show a slight advantage over traditional encoder models if they are properly fine-tuned for the task. This trend suggests that, for skill relevance tasks, embedding models built on LLMs and enhanced with advanced data augmentation techniques can offer additional benefits over simpler approaches.

\section{Conclusions}

TalentCLEF 2025 has established the first community evaluation framework for NLP applications in Human Capital Management, addressing a gap in this evolving field. The initiative attracted significant participation, with 76 registered teams contributing with more than 280 submissions in two tasks useful for real-world talent management challenges such as job matching and skill-based job matching. 

The technical results challenge some prevailing assumptions about the performance of the model. Encoder-based models fine-tuned with contrastive learning techniques consistently outperformed larger decoder-based embedding system. Even in cases like Task B, where a decoder-based model showed better performance, its size, cost, and computational requirements arguably may not justify their use in a real scenario. Still, the most successful systems took advantage of autoregressive language models to support tasks such as data augmentation, leading to better models as a result of training them with more diverse data.

%Even in cases like Task B, where a decoder-based model achieved better performance, their size and computational requirements may not justify their use in a real scenario. However, the most successful systems leveraged the capabilities of autoregressive language models for auxiliary tasks such as data augmentation, resulting in better models, as they are trained on more diverse data.

From an evaluation perspective,  the systems showed strong performance in both languages and in terms of fairness. In fact, some approaches generated genuinely fair predictions, producing exactly the same results for male and female gendered job titles.

Beyond the technical contribution of the submitted systems, TalentCLEF has established something crucial for this area which is highly dependent on language technologies: shared, open, and publicly available evaluation methods and datasets that capture the true complexity of real-world data.  The benchmarks created in this challenge not only support ongoing research, but also help guide the responsible use of language technologies in areas that deeply affect people's lives, such as their professional careers.

\section{Acknowledgments}
% Recognition of the annotatiors that did a good work
We would like to acknowledge Aurelia Cañete, Laura Bruno, Mariano Cabrera, Martín Gabriel Chaine Pasco, Georgiana Zarnescu, and Thomas Krucky for their collaboration in the annotation of the tasks data.

\bibliographystyle{splncs04}  % El estilo que usa Springer LNCS
\bibliography{bibliography}

\begin{thebibliography}{10}
\providecommand{\url}[1]{\texttt{#1}}
\providecommand{\urlprefix}{URL }
\providecommand{\doi}[1]{https://doi.org/#1}

\bibitem{ali2025enhancingjob}
Ali, M.: {Enhancing Job-Skill Matching with LLM-Driven Data Augmentation and Fine-Tuned Bi-Encoders}. In: {CLEF (Working Notes)} (2025)

\bibitem{barakat2025alexunlp}
Barakat, R., Mokhtar, O., Torki, M., Elmakky, N.: {AlexU-NLP at TalentCLEF 2025: Curriculum-Driven Hybrid Retrieval for Multilingual Job Title Matching}. In: {CLEF (Working Notes)} (2025)

\bibitem{bhangale2025finetuned}
Bhangale, C., Gabhane, P., Kumar~M, A.: {Fine-Tuned Sentence Transformer for Multilingual Job Title Matching}. In: {CLEF (Working Notes)} (2025)

\bibitem{bogers2024fourth}
Bogers, T., Graus, D., Kaya, M., Johnson, C., Decorte, J.J., De~Bie, T.: Fourth workshop on recommender systems for human resources (recsys in hr 2024). In: Proceedings of the 18th ACM Conference on Recommender Systems. p. 1222–1226. RecSys '24, Association for Computing Machinery, New York, NY, USA (2024). \doi{10.1145/3640457.3687109}, \url{https://doi.org/10.1145/3640457.3687109}

\bibitem{brikman2025multilingualjob}
Brikman, A., Sana, M., Ruegger, H.: {Multilingual Job Title Matching with MPNet-Based Sentence Transformers}. In: {CLEF (Working Notes)} (2025)

\bibitem{chen2024m3}
Chen, J., Xiao, S., Zhang, P., Luo, K., Lian, D., Liu, Z.: M3-embedding: Multi-linguality, multi-functionality, multi-granularity text embeddings through self-knowledge distillation. In: Ku, L., Martins, A., Srikumar, V. (eds.) Findings of the Association for Computational Linguistics, {ACL} 2024, Bangkok, Thailand and virtual meeting, August 11-16, 2024. pp. 2318--2335. Association for Computational Linguistics (2024). \doi{10.18653/V1/2024.FINDINGS-ACL.137}, \url{https://doi.org/10.18653/v1/2024.findings-acl.137}

\bibitem{choi2024linq}
Choi, C., Kim, J., Lee, S., Kwon, J., Gu, S., Kim, Y., Cho, M., Sohn, J.: Linq-embed-mistral technical report. CoRR  \textbf{abs/2412.03223} (2024). \doi{10.48550/ARXIV.2412.03223}, \url{https://doi.org/10.48550/arXiv.2412.03223}

\bibitem{decorte2025techwolf}
Decorte, J.J., De~Lange, M., Van~Hautte, J.: {TechWolf at TalentCLEF 2025: Multilingual JobBERT-V2 for Cross-Lingual Job Title Matching}. In: {CLEF (Working Notes)} (2025)

\bibitem{decorte2021jobbert}
Decorte, J., Hautte, J.V., Demeester, T., Develder, C.: Jobbert: Understanding job titles through skills. CoRR  \textbf{abs/2109.09605} (2021), \url{https://arxiv.org/abs/2109.09605}

\bibitem{deniz2024combined}
Deniz, D., Retyk, F., Garc{\'{\i}}a{-}Sardi{\~{n}}a, L., Fabregat, H., Gasc{\'{o}}, L., Zbib, R.: Combined unsupervised and contrastive learning for multilingual job recommendation. In: Kaya, M., Bogers, T., Graus, D., Johnson, C., Decorte, J., Bie, T.D. (eds.) Proceedings of the 4th Workshop on Recommender Systems for Human Resources (RecSys-in-HR 2024) co-located with the 18th {ACM} Conference on Recommender Systems (RecSys 2024), Bari, Italy, 14th-18th October 2024. {CEUR} Workshop Proceedings, vol.~3788. CEUR-WS.org (2024), \url{https://ceur-ws.org/Vol-3788/RecSysHR2024-paper\_3.pdf}

\bibitem{di2023future}
{Di Battista, Attilio and Grayling, Sam and Hasselaar, Elselot and Leopold, Till and Li, Ricky and Rayner, Mark and Zahidi, Saadia}: {Future of jobs report 2023}. In: {World Economic Forum, Geneva, Switzerland. https://www. weforum. org/reports/the-future-of-jobs-report-2023}. World Economic Forum (2023)

\bibitem{grattafiori2024llama}
Dubey, A., Jauhri, A., Pandey, A., Kadian, A., Al{-}Dahle, A., Letman, A., Mathur, A., Schelten, A., Yang, A., Fan, A., Goyal, A., Hartshorn, A., Yang, A., Mitra, A., Sravankumar, A., Korenev, A., Hinsvark, A., Rao, A., Zhang, A., Rodriguez, A., Gregerson, A., Spataru, A., Rozi{\`{e}}re, B., Biron, B., Tang, B., Chern, B., Caucheteux, C., Nayak, C., Bi, C., Marra, C., McConnell, C., Keller, C., Touret, C., Wu, C., Wong, C., Ferrer, C.C., Nikolaidis, C., Allonsius, D., Song, D., Pintz, D., Livshits, D., Esiobu, D., Choudhary, D., Mahajan, D., Garcia{-}Olano, D., Perino, D., Hupkes, D., Lakomkin, E., AlBadawy, E., Lobanova, E., Dinan, E., Smith, E.M., Radenovic, F., Zhang, F., Synnaeve, G., Lee, G., Anderson, G.L., Nail, G., Mialon, G., Pang, G., Cucurell, G., Nguyen, H., Korevaar, H., Xu, H., Touvron, H., Zarov, I., Ibarra, I.A., Kloumann, I.M., Misra, I., Evtimov, I., Copet, J., Lee, J., Geffert, J., Vranes, J., Park, J., Mahadeokar, J., Shah, J., van~der Linde, J., Billock, J., Hong, J., Lee, J., Fu, J.,
  Chi, J., Huang, J., Liu, J., Wang, J., Yu, J., Bitton, J., Spisak, J., Park, J., Rocca, J., Johnstun, J., Saxe, J., Jia, J., Alwala, K.V., Upasani, K., Plawiak, K., Li, K., Heafield, K., Stone, K., et~al.: The llama 3 herd of models. CoRR  \textbf{abs/2407.21783} (2024). \doi{10.48550/ARXIV.2407.21783}, \url{https://doi.org/10.48550/arXiv.2407.21783}

\bibitem{fabregat2024inductive}
Fabregat, H., Poves, R., Alvarez, L.L., Retyk, F., Garc{\'{\i}}a{-}Sardi{\~{n}}a, L., Zbib, R.: Inductive graph neural network for job-skill framework analysis. Procesamiento del Lenguaje Natural  \textbf{73},  83--94 (2024), \url{http://journal.sepln.org/sepln/ojs/ojs/index.php/pln/article/view/6602}

\bibitem{gasco2025talentclef}
Gasco, L., Fabregat, H., Garc{\'\i}a-Sardi{\~n}a, L., Deniz, D., Rodrigo, A., Estrella, P., Zbib, R.: {TalentCLEF at CLEF2025: Skill and Job Title Intelligence for Human Capital Management}. In: European Conference on Information Retrieval. pp. 479--486. Springer (2025)

\bibitem{giabelli2021skills2job}
Giabelli, A., Malandri, L., Mercorio, F., Mezzanzanica, M., Seveso, A.: Skills2job: {A} recommender system that encodes job offer embeddings on graph databases. Applied Soft Computing  \textbf{101},  107049 (2021). \doi{10.1016/J.ASOC.2020.107049}, \url{https://doi.org/10.1016/j.asoc.2020.107049}

\bibitem{ho2025ntteam}
Ho, T.N., Ho, T.T.T., Dang, V.T.: {NT Team at Multilingual Job Title Matching Task A: Job Matching via Large Language Model-Based Description Generation and Retrieval}. In: {CLEF (Working Notes)} (2025)

\bibitem{hruschka2024proceedings}
Hruschka, E., Lake, T., Otani, N., Mitchell, T.: Proceedings of the first workshop on natural language processing for human resources (nlp4hr 2024). In: Proceedings of the First Workshop on Natural Language Processing for Human Resources (NLP4HR 2024). "Association for Computational Linguistics" (2024), \url{https://aclanthology.org/2024.nlp4hr-1.0/}

\bibitem{ai4hrpes2023}
Kang, B., De~Bie, T., Sebag, M., Largeron, C.: Ai for human resources and people analytics workshop program, ecml pkdd 2023. \url{https://ai4hrpes.github.io/ecmlpkdd2023/program/}, accessed: 2024-10-17

\bibitem{lavi2021consultantbert}
Lavi, D., Medentsiy, V., Graus, D.: consultantbert: Fine-tuned siamese sentence-bert for matching jobs and job seekers. In: Kaya, M., Bogers, T., Graus, D., Verbert, K., Guti{\'{e}}rrez, F. (eds.) Proceedings of the Workshop on Recommender Systems for Human Resources (RecSys in {HR} 2021) co-located with the 15th {ACM} Conference on Recommender Systems (RecSys 2021), Amsterdam, The Netherlands, 27th September - 1st October 2021. {CEUR} Workshop Proceedings, vol.~2967. CEUR-WS.org (2021), \url{https://ceur-ws.org/Vol-2967/paper\_8.pdf}

\bibitem{li2023towards}
Li, Z., Zhang, X., Zhang, Y., Long, D., Xie, P., Zhang, M.: Towards general text embeddings with multi-stage contrastive learning. CoRR  \textbf{abs/2308.03281} (2023). \doi{10.48550/ARXIV.2308.03281}, \url{https://doi.org/10.48550/arXiv.2308.03281}

\bibitem{linkedin2025workchange}
{LinkedIn}: Work change report: Ai is coming to work (January 2025), \url{https://economicgraph.linkedin.com/content/dam/me/economicgraph/en-us/PDF/Work-Change-Report.pdf}, accessed: 2025-05-18

\bibitem{llamas2025cotecmarutb}
Llamas, J., Puertas, E., Serrano, J., Martinez, J.: {COTECMAR–UTB at TalentCLEF 2025: Linking Job Titles and ESCO Skills with Sentence Transformer Embeddings}. In: {CLEF (Working Notes)} (2025)

\bibitem{manpower2024talentshortage}
{ManpowerGroup}: 2024 global talent shortage (2024), \url{https://go.manpowergroup.com}, accessed: 2025-05-19

\bibitem{miranda2023overview}
Miranda-Escalada, A., Mehryary, F., Luoma, J., Estrada-Zavala, D., Gasco, L., Pyysalo, S., Valencia, A., Krallinger, M.: Overview of drugprot task at biocreative vii: data and methods for large-scale text mining and knowledge graph generation of heterogenous chemical--protein relations. Database  \textbf{2023},  1--24 (2023). \doi{10.1093/database/baad080}, \url{https://doi.org/10.1093/database/baad080}

\bibitem{moreno2025verbanex}
Moreno~Novoa, M., Mart\'inez-Santos, J.C., Serrano, J., Puertas, E.: {VerbaNex at TalentCLEF2025: Semantic Matching of Multilingual Job Titles through a Framework Integrating ESCO Taxonomy}. In: {CLEF (Working Notes)} (2025)

\bibitem{nentidis2022overview}
Nentidis, A., Katsimpras, G., Vandorou, E., Krithara, A., Miranda-Escalada, A., Gasco, L., Krallinger, M., Paliouras, G.: Overview of bioasq 2022: the tenth bioasq challenge on large-scale biomedical semantic indexing and question answering. In: Experimental IR Meets Multilinguality, Multimodality, and Interaction. pp. 337--361. Springer International Publishing, Cham (2022)

\bibitem{oord2018representation}
van~den Oord, A., Li, Y., Vinyals, O.: Representation learning with contrastive predictive coding. CoRR  \textbf{abs/1807.03748} (2018), \url{http://arxiv.org/abs/1807.03748}

\bibitem{retyk2023r}
Retyk, F., Fabregat, H., Aizpuru, J., Taglio, M., Zbib, R.: R{\'{e}}sum{\'{e}} parsing as hierarchical sequence labeling: An empirical study. In: Kaya, M., Bogers, T., Graus, D., Johnson, C., Decorte, J. (eds.) Proceedings of the 3rd Workshop on Recommender Systems for Human Resources (RecSys in {HR} 2023) co-located with the 17th {ACM} Conference on Recommender Systems (RecSys 2023), Singapore, Singapore, 18th-22nd September 2023. {CEUR} Workshop Proceedings, vol.~3490. CEUR-WS.org (2023), \url{https://ceur-ws.org/Vol-3490/RecSysHR2023-paper\_10.pdf}

\bibitem{retyk2024melo}
Retyk, F., Gasc{\'{o}}, L., Carrino, C.P., Deniz, D., Zbib, R.: {MELO:} an evaluation benchmark for multilingual entity linking of occupations. In: Kaya, M., Bogers, T., Graus, D., Johnson, C., Decorte, J., Bie, T.D. (eds.) Proceedings of the 4th Workshop on Recommender Systems for Human Resources (RecSys-in-HR 2024) co-located with the 18th {ACM} Conference on Recommender Systems (RecSys 2024), Bari, Italy, 14th-18th October 2024. {CEUR} Workshop Proceedings, vol.~3788. CEUR-WS.org (2024), \url{https://ceur-ws.org/Vol-3788/RecSysHR2024-paper\_2.pdf}

\bibitem{team2024gemma}
Rivi{\`{e}}re, M., Pathak, S., Sessa, P.G., Hardin, C., Bhupatiraju, S., Hussenot, L., Mesnard, T., Shahriari, B., Ram{\'{e}}, A., Ferret, J., Liu, P., Tafti, P., Friesen, A., Casbon, M., Ramos, S., Kumar, R., Lan, C.L., Jerome, S., Tsitsulin, A., Vieillard, N., Stanczyk, P., Girgin, S., Momchev, N., Hoffman, M., Thakoor, S., Grill, J., Neyshabur, B., Bachem, O., Walton, A., Severyn, A., Parrish, A., Ahmad, A., Hutchison, A., Abdagic, A., Carl, A., Shen, A., Brock, A., Coenen, A., Laforge, A., Paterson, A., Bastian, B., Piot, B., Wu, B., Royal, B., Chen, C., Kumar, C., Perry, C., Welty, C., Choquette{-}Choo, C.A., Sinopalnikov, D., Weinberger, D., Vijaykumar, D., Rogozinska, D., Herbison, D., Bandy, E., Wang, E., Noland, E., Moreira, E., Senter, E., Eltyshev, E., Visin, F., Rasskin, G., Wei, G., Cameron, G., Martins, G., Hashemi, H., Klimczak{-}Plucinska, H., Batra, H., Dhand, H., Nardini, I., Mein, J., Zhou, J., Svensson, J., Stanway, J., Chan, J., Zhou, J.P., Carrasqueira, J., Iljazi, J., Becker, J.,
  Fernandez, J., van Amersfoort, J., Gordon, J., Lipschultz, J., Newlan, J., Ji, J., Mohamed, K., Badola, K., Black, K., Millican, K., McDonell, K., Nguyen, K., Sodhia, K., Greene, K., Sj{\"{o}}sund, L.L., Usui, L., Sifre, L., Heuermann, L., Lago, L., McNealus, L.: Gemma 2: Improving open language models at a practical size. CoRR  \textbf{abs/2408.00118} (2024). \doi{10.48550/ARXIV.2408.00118}, \url{https://doi.org/10.48550/arXiv.2408.00118}

\bibitem{rodriguez2025twostage}
Rodr\'iguez, M., Perez-de Viñaspre, O., Perez, N.: {A Two-Stage Multilingual Job Title Matching System: Combining Expert Knowledge and LLM-based Ranking}. In: {CLEF (Working Notes)} (2025)

\bibitem{rodriguezvidal2025udiiupm}
Rodr\'iguez-Vidal, J., L\'opez-Vargas, A., Vigara~Gallego, P.M., Del~\'Alamo, F.J., Garc\'ia-Beltr\'an, A.: {UDII-UPM at TalentCLEF 2025: Task A-Multilingual Job Title Matching}. In: {CLEF (Working Notes)} (2025)

\bibitem{senger2024deep}
Senger, E., Zhang, M., van~der Goot, R., Plank, B.: Deep learning-based computational job market analysis: A survey on skill extraction and classification from job postings. In: Proceedings of the First Workshop on Natural Language Processing for Human Resources (NLP4HR 2024). pp. 1--15. "Association for Computational Linguistics", "St. Julian{'}s, Malta" (2024), \url{https://aclanthology.org/2024.nlp4hr-1.1/}

\bibitem{smit2020future}
{Smit, Sven and Tacke, Tilman and Lund, Susan and Manyika, James and Thiel, Lea}: {The future of work in Europe: Automation, workforce transitions, and the shifting geography of employment}. Tech. rep., {McKinsey \& Company} (2020)

\bibitem{solatorio2024gistembed}
Solatorio, A.V.: Gistembed: Guided in-sample selection of training negatives for text embedding fine-tuning. CoRR  \textbf{abs/2402.16829} (2024). \doi{10.48550/ARXIV.2402.16829}, \url{https://doi.org/10.48550/arXiv.2402.16829}

\bibitem{soviany2022curriculum}
Soviany, P., Ionescu, R.T., Rota, P., Sebe, N.: Curriculum learning: {A} survey. International Journal of Computer Vision  \textbf{130}(6),  1526--1565 (2022). \doi{10.1007/S11263-022-01611-X}, \url{https://doi.org/10.1007/s11263-022-01611-x}

\bibitem{tahmid2024qwen2}
Tahmid, S., Sarker, S.: Qwen2.5-32b: Leveraging self-consistent tool-integrated reasoning for bengali mathematical olympiad problem solving. CoRR  \textbf{abs/2411.05934} (2024). \doi{10.48550/ARXIV.2411.05934}, \url{https://doi.org/10.48550/arXiv.2411.05934}

\bibitem{tejera2025hulatuc3m}
Tejera~Villar, A., Segura~Bedmar, I.: {HULAT-UC3M at TalentCLEF: Artificial Intelligence and Natural Language Processing applied to HR Management}. In: {CLEF (Working Notes)} (2025)

\bibitem{uddin2025enhancing}
Uddin, A., Nizami, M.H., Salani, M.T., Saeed, A.: {Enhancing Human Capital Management: AI Techniques for Candidate Matching and Skill Extraction}. In: {CLEF (Working Notes)} (2025)

\bibitem{vachharajani2025pjmathematician}
Vachharajani, P.: {pjmathematician at TalentCLEF 2025: Enhancing Job Title and Skill Matching with GISTEmbed and LLM-Augmented Data}. In: {CLEF (Working Notes)} (2025)

\bibitem{vazquezgarcia2025beyondtitles}
V\'azquez~Garc\'ia, I.X., Sedano~Puente, R., Gonz\'alez~Gonz\'alez, S., Sedano~Franco, J.: {Beyond Titles: Semantic Matching of Jobs and Skills Using LLMs and S-BERT}. In: {CLEF (Working Notes)} (2025)

\bibitem{wang2024multilingual}
Wang, L., Yang, N., Huang, X., Yang, L., Majumder, R., Wei, F.: Multilingual {E5} text embeddings: {A} technical report. CoRR  \textbf{abs/2402.05672} (2024). \doi{10.48550/ARXIV.2402.05672}, \url{https://doi.org/10.48550/arXiv.2402.05672}

\bibitem{wmz10:acmtois}
Webber, W., Moffat, A., Zobel, J.: A similarity measure for indefinite rankings. ACM Transactions on Information Systems  (2010)

\bibitem{zbib2022learning}
Zbib, R., Alvarez, L.L., Retyk, F., Poves, R., Aizpuru, J., Fabregat, H., Simkus, V., Casademont, E.G.: Learning job titles similarity from noisy skill labels. CoRR  \textbf{abs/2207.00494} (2022). \doi{10.48550/ARXIV.2207.00494}, \url{https://doi.org/10.48550/arXiv.2207.00494}

\bibitem{zhang2025nlpnorth}
Zhang, M., van~der Goot, R.: {NLPnorth @ TalentCLEF 2025: Discriminative vs. Contrastive vs. Prompting for Job Title and Job-Skill Matching}. In: {CLEF (Working Notes)} (2025)

\bibitem{zhang2023escoxlm}
Zhang, M., van~der Goot, R., Plank, B.: {ESCOXLM-R:} multilingual taxonomy-driven pre-training for the job market domain. In: Rogers, A., Boyd{-}Graber, J.L., Okazaki, N. (eds.) Proceedings of the 61st Annual Meeting of the Association for Computational Linguistics (Volume 1: Long Papers), {ACL} 2023, Toronto, Canada, July 9-14, 2023. pp. 11871--11890. Association for Computational Linguistics (2023). \doi{10.18653/V1/2023.ACL-LONG.662}, \url{https://doi.org/10.18653/v1/2023.acl-long.662}

\bibitem{zhang2022skill}
Zhang, M., Jensen, K.N., van~der Goot, R., Plank, B.: Skill extraction from job postings using weak supervision. In: Kaya, M., Bogers, T., Graus, D., Mesbah, S., Johnson, C., Guti{\'{e}}rrez, F. (eds.) Proceedings of the 2nd Workshop on Recommender Systems for Human Resources (RecSys-in-HR 2022) co-located with the 16th {ACM} Conference on Recommender Systems (RecSys 2022), Seattle, USA, 18th-23rd September 2022. {CEUR} Workshop Proceedings, vol.~3218. CEUR-WS.org (2022), \url{https://ceur-ws.org/Vol-3218/RecSysHR2022-paper\_10.pdf}

\bibitem{zhang2022skillspan}
Zhang, M., Jensen, K.N., Sonniks, S.D., Plank, B.: Skillspan: Hard and soft skill extraction from english job postings. In: Carpuat, M., de~Marneffe, M., Ru{\'{\i}}z, I.V.M. (eds.) Proceedings of the 2022 Conference of the North American Chapter of the Association for Computational Linguistics: Human Language Technologies, {NAACL} 2022, Seattle, WA, United States, July 10-15, 2022. pp. 4962--4984. Association for Computational Linguistics (2022). \doi{10.18653/V1/2022.NAACL-MAIN.366}, \url{https://doi.org/10.18653/v1/2022.naacl-main.366}

\bibitem{zhu20245th}
Zhu, H., Ge, Y., Xiong, H., Lim, E.P.: The 5th international workshop on talent and management computing (tmc'2024). In: Proceedings of the 30th ACM SIGKDD Conference on Knowledge Discovery and Data Mining. p. 6759–6760. KDD '24, Association for Computing Machinery, New York, NY, USA (2024). \doi{10.1145/3637528.3671479}, \url{https://doi.org/10.1145/3637528.3671479}

\end{thebibliography}

\end{document}